\pdfoutput=1

\documentclass[11pt]{article}
\usepackage[table]{xcolor}
\definecolor{cvprblue}{RGB}{0,114,178} 
\usepackage[most]{tcolorbox}
\usepackage[dvipsnames]{xcolor}
\usepackage{makecell}
\usepackage{booktabs}
\usepackage{siunitx}
\usepackage{amssymb}
\usepackage{multirow}
\usepackage[final]{acl}
\usepackage[ruled,vlined]{algorithm2e}
\usepackage{times}
\usepackage{latexsym}
\usepackage{amsmath}
\usepackage{graphicx}  
\usepackage[T1]{fontenc}

\usepackage[utf8]{inputenc}

\usepackage{microtype}

\usepackage{inconsolata}

\usepackage{graphicx}
\usepackage{hyperref}

\title{From Long to Lean: Performance-aware and Adaptive Chain-of-Thought Compression via Multi-round Refinement}

\author{
  Jianzhi Yan\textsuperscript{1,2},
  Le Liu\textsuperscript{1,2},
  Youcheng Pan\textsuperscript{2}\footnotemark[1],
  Shiwei Chen\textsuperscript{1,2} \\
  \textbf{Zike Yuan\textsuperscript{1,2}},
  \textbf{Yang Xiang\textsuperscript{2,3}}\thanks{Corresponding authors.},
  \textbf{Buzhou Tang\textsuperscript{1,2}}\footnotemark[1] \\
  \textsuperscript{1}Harbin Institute of Technology, Shenzhen, China \\
  \textsuperscript{2}Pengcheng Laboratory, Shenzhen, China \\
  \textsuperscript{3}Shaoguan Research Institute of Data Industry, China \\
  \texttt{\{yanjzh, liul07, panych, chenshw, yuanzk, xiangy\}@pcl.ac.cn} \\
  \texttt{tangbuzhou@gmail.com}
}

\begin{document}
\maketitle
\begin{abstract}
Chain-of-Thought (CoT) reasoning improves performance on complex tasks but introduces significant inference latency due to its verbosity. In this work, we propose Multiround Adaptive Chain-of-Thought Compression (\textbf{MACC}), a framework that leverages the \textit{token elasticity phenomenon}—where overly small token budgets may paradoxically increase output length—to progressively compress CoTs via multiround refinement. This adaptive strategy allows MACC to dynamically determine the optimal compression depth for each input. Our method achieves an average accuracy improvement of 5.6$\%$ over state-of-the-art baselines, while also reducing CoT length by an average of 47 tokens and significantly lowering latency. Furthermore, we show that \textbf{test-time performance}—accuracy and token length—can be reliably predicted using interpretable features like perplexity and compression rate \textbf{on training set}. Evaluated across different models, our method enables efficient model selection and forecasting without repeated fine-tuning, demonstrating that CoT compression is both effective and predictable. Our code will be released in \url{https://github.com/Leon221220/MACC}.
\end{abstract}

\section{Introduction}

Chain-of-Thought (CoT) reasoning significantly enhances the performance of large language models (LLMs) on complex tasks by decomposing questions into intermediate steps and reasoning sequentially \cite{nye2021workscratchpadsintermediatecomputation, wei2023chainofthoughtpromptingelicitsreasoning, kojima2023largelanguagemodelszeroshot}.
Recent models such as OpenAI-o1 \cite{openai2024openaio1card} and DeepSeek-R1 \cite{deepseekai2025deepseekr1incentivizingreasoningcapability} demonstrate that Test-Time Scaling (TTS)—increasing CoT length during inference—can further boost reasoning accuracy \cite{snell2024scalingllmtesttimecompute, tian2025thinktwiceenhancingllm, yang2025thinkingoptimalscalingtesttimecompute}.
Nevertheless, longer CoTs substantially increase inference latency and memory usage due to larger key-value caches and the quadratic complexity of attention for sequence length \cite{dao2022flashattentionfastmemoryefficientexact, liu2024retrievalattentionacceleratinglongcontextllm, aisagescribe2025kv}. To address the inefficiency of lengthy CoT reasoning, recent work has proposed a range of compression strategies. Token-level methods (TokenSkip \cite{han2025tokenbudgetawarellmreasoning}, C3oT \cite{kang2024c3otgeneratingshorterchainofthought}) prune redundant steps and use fine-tuning to preserve performance under compression \cite{jang2024verbosityawarerationalereductioneffective, liu2024languagemodelslearnskip, cui2025stepwiseperplexityguidedrefinementefficient, shen2025efficientreasoninghiddenthinking, yu2024distilling21}. Prompt-based approaches (CoD \cite{xu2025chaindraftthinkingfaster}, SoT \cite{aytes2025sketchofthoughtefficientllmreasoning}, TALE-EP \cite{han2025tokenbudgetawarellmreasoning}) guide concise reasoning via routing prompts, minimal templates, or difficulty-aware designs \cite{yan2025inftythinkbreakinglengthlimits, zhang2025lightthinkerthinkingstepbystepcompression, sui2025metareasonerdynamicguidanceoptimized}. Reward-based methods (O1-Pruner \cite{luo2025o1prunerlengthharmonizingfinetuningo1like}, DAST \cite{shen2025dastdifficultyadaptiveslowthinkinglarge}, IBPO \cite{xu2025chaindraftthinkingfaster}) optimize reasoning length via reinforcement learning or preference modeling \cite{qu2025optimizingtesttimecomputemeta, yeo2025demystifyinglongchainofthoughtreasoning, chen2025think23overthinkingo1like}.

\begin{figure}[!t]
\centering
\includegraphics[scale=0.15]{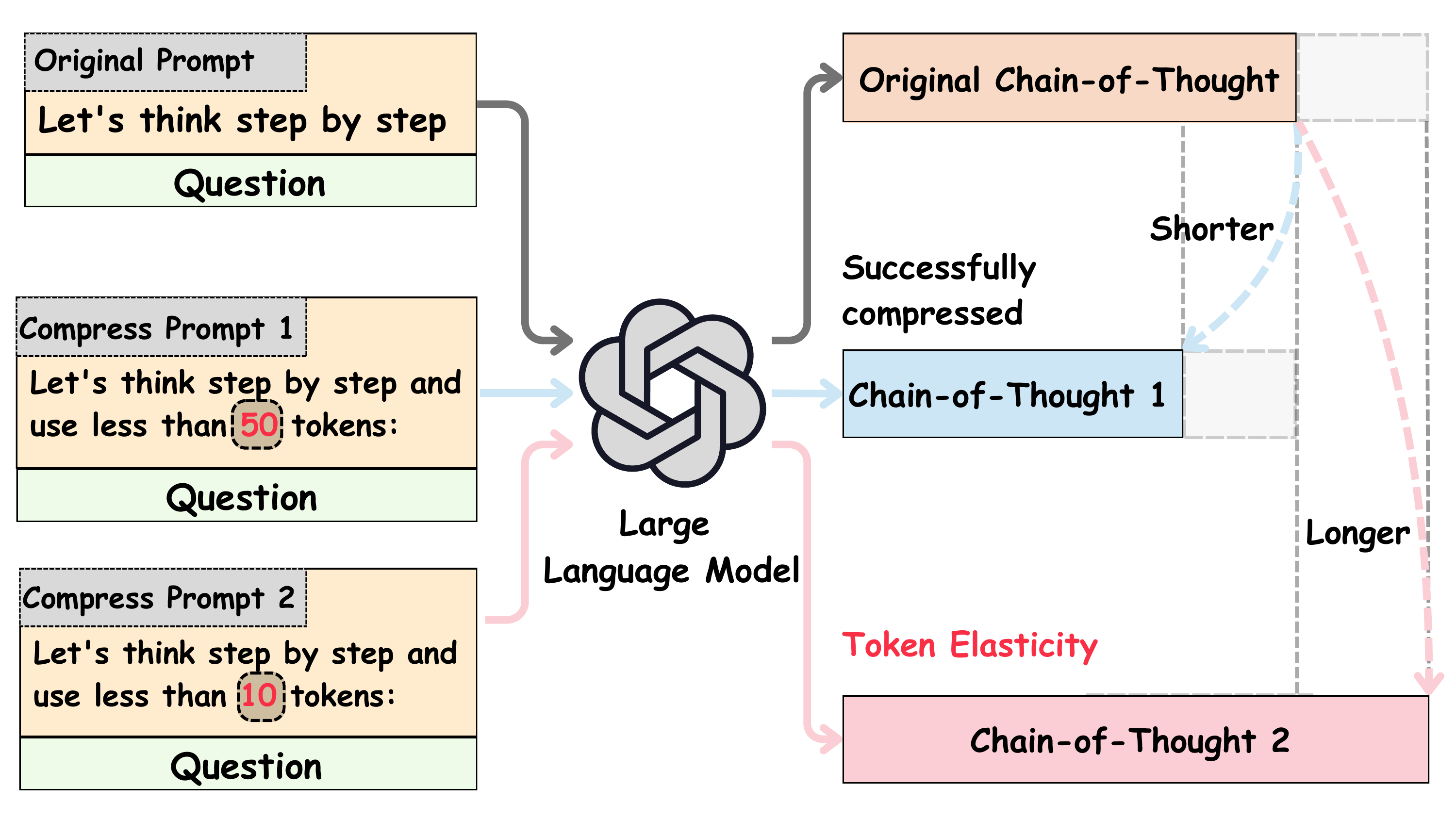}
\caption{
\textbf{Visualization of the \textit{Token Elasticity} phenomenon}. As the prompt-specified token budget decreases, the actual token cost initially declines but eventually rebounds when the budget becomes too small.}
\label{fig:token_elasticity}
\end{figure}

\par However, prior approaches \textbf{lack fine-grained adaptability in managing the trade-off between compression and accuracy across diverse reasoning inputs}.
Pruning- and prompt-based methods typically apply uniform compression, ignoring input-specific reasoning complexity, while reward-driven strategies optimize global preferences without instance-level control. For example, TokenSkip performs static token pruning and often degrades performance under tight budgets \cite{xia2025tokenskipcontrollablechainofthoughtcompression}; CoD uses fixed prompts without controlling reasoning depth per instance \cite{xu2025chaindraftthinkingfaster}; and TALE, though budget-aware, compresses in a single pass without adapting to input difficulty \cite{han2025tokenbudgetawarellmreasoning}. These methods lack the adaptive refinement needed to balance efficiency and accuracy in a controllable, input-sensitive manner.

\par To address these issues, we proposed Multiround Adaptive Chain-of-Thought Compression, a framework grounded in the observed phenomenon of \textit{token elasticity}-as shown in Figure~\ref{fig:token_elasticity}—where overly aggressive compression may paradoxically increase token usage due to degraded generation quality. Our framework consists of three main components: (1) Chain-of-thought generation, (2) Multi-round progressive compression, and (3) Multitask fine-tuning. Given a question, we first prompt a model to generate a full reasoning trace, which is then progressively compressed through multiple rounds using compressor models. Each round removes redundant or verbose steps while preserving essential information, with dynamic control over compression ratios to adapt granularity. The final compressed CoTs are used to fine-tune models for efficient inference. Moreover, we proposed \textbf{\textit{Performance Estimation Hypothesis}: test-time performance of the compressed CoT can be estimated \emph{before fine-tuning}, based solely on a small set of interpretable features derived from the training set}—including compression rate, perplexity, original model training set accuracy, and average training set CoT length. We train lightweight regression models to predict both the downstream accuracy and token efficiency of the target model on the test set, enabling early-stage compression strategy selection without costly retraining. This predictive capacity makes our framework both efficient and performance-aware.

To sum up, our key contributions are:
\par 1. We propose MACC, a multi-round compression framework that adaptively shortens reasoning chains while preserving essential information.
\par 2. MACC achieves 5.6$\%$ higher accuracy, reduces reasoning by 47 tokens on average, and lowers latency, while supporting efficient model selection via interpretable metrics.
\par 3. We propose \textbf{\textit{Performance Estimation Hypothesis}} and demonstrate that fine-tuned performance can be predicted from interpretable features on the training set, enabling efficient strategy selection.

\section{Related Work}

\subsection{LLM Reasoning and Token Cost}
Recent advances in LLM reasoning techniques, particularly CoT prompting and its extensions such as self-consistency and tree-structured reasoning, have significantly enhanced complex problem-solving capabilities \cite{wei2023chainofthoughtpromptingelicitsreasoning, wang2023selfconsistencyimproveschainthought, yao2023treethoughtsdeliberateproblem, zhou2023leasttomostpromptingenablescomplex}. A variety of techniques have been proposed to enhance LLM reasoning. \citet{chen2024languagemodelshiddenreasoners} frame reasoning as latent distribution sampling optimized via variational methods, while \citet{ho2023largelanguagemodelsreasoning} leverages LLMs as reasoning teachers to distill knowledge into smaller models. But at the cost of substantially increased token consumption and computational overhead \cite{wang2024reasoningtokeneconomiesbudgetaware, chiang2024overreasoningredundantcalculationlarge, bhargava2024whatsmagicwordcontrol}. To improve efficiency, \citealt{li-etal-2021-addressing-semantic} propose a multi-hop filtering method to discard irrelevant reasoning, but it is limited to traditional neural networks and does not generalize to LLMs. \citet{zheng2023responselengthperceptionsequence} enhance inference speed via response length prediction and scheduling, yet their method operates only at the scheduling level without reducing token usage. \citet{hao2024traininglargelanguagemodels} lowers token cost by replacing decoded text with continuous latent tokens.

\begin{figure*}[!ht]
  \includegraphics[scale=0.31]{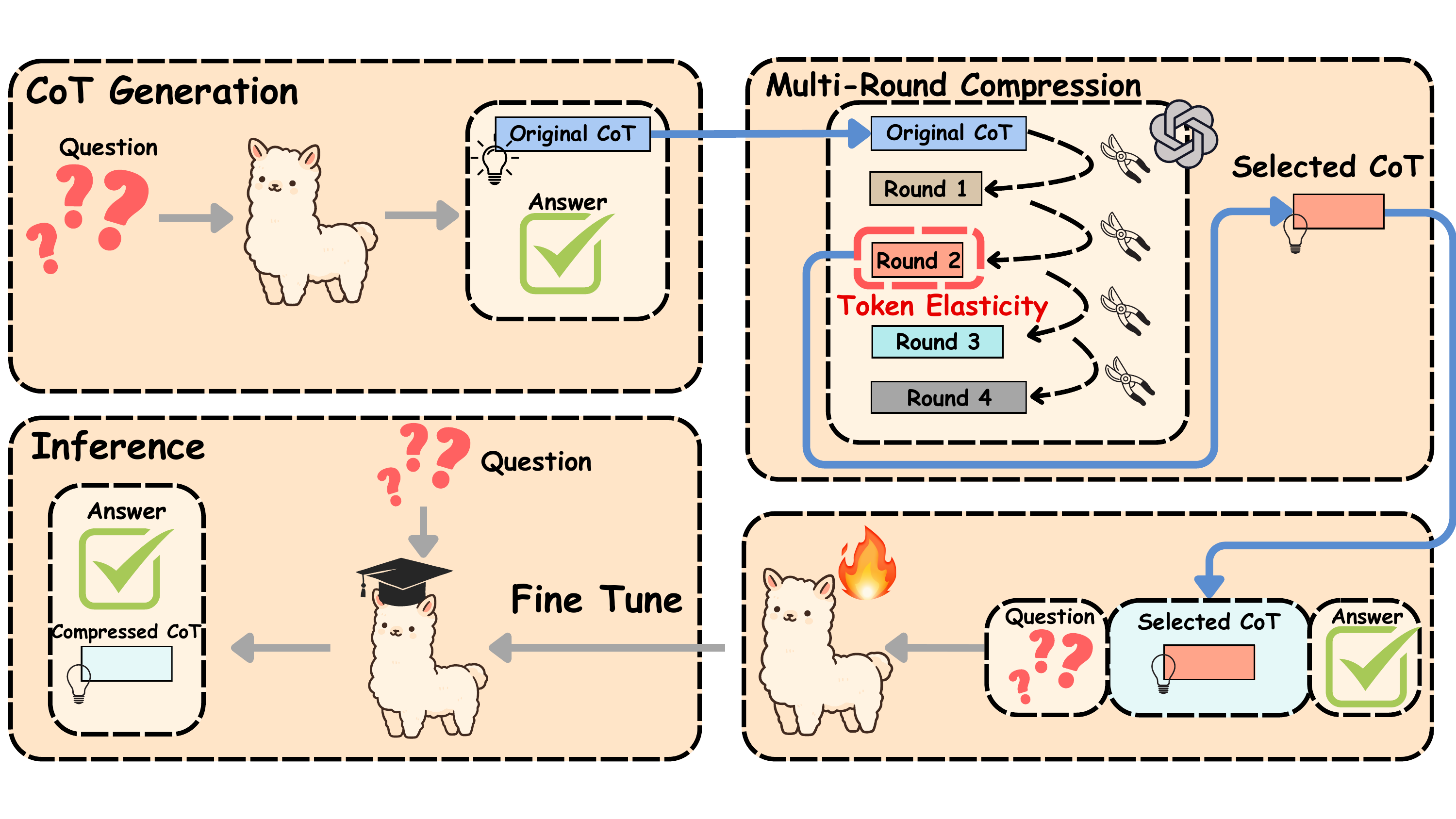}
  \caption{\textbf{Overview of MACC framework}. Given an input question, model first generates a full reasoning trace (CoT). The CoT is then progressively compressed through multiple rounds using a compressor model to remove redundancy while retaining essential reasoning content. The resulting compressed CoTs are used to fine-tune a smaller target model for efficient inference.}
  \label{fig:detailed}
\end{figure*}

\subsection{Chain-of-Thought Compression}

To improve LLM inference efficiency, recent work explores compressing CoT reasoning while preserving answer correctness. These approaches can be broadly categorized into three paradigms \cite{liu2025efficientinferencelargereasoning, qu2025surveyefficientreasoninglarge}. First, Token-level compression methods, such as TokenSkip \cite{han2025tokenbudgetawarellmreasoning} and C3oT \cite{kang2024c3otgeneratingshorterchainofthought}, prune redundant tokens or steps and use supervised fine-tuning to maintain accuracy under varying compression ratios \cite{jang2024verbosityawarerationalereductioneffective, liu2024languagemodelslearnskip}. Second, Prompt design and sketch-based approaches, including CoD \cite{xu2025chaindraftthinkingfaster}, SoT \cite{aytes2025sketchofthoughtefficientllmreasoning}, and TALE-EP \cite{han2025tokenbudgetawarellmreasoning}, guide concise reasoning using routing prompts, minimalist structures, or token-aware templates \cite{yan2025inftythinkbreakinglengthlimits, zhang2025lightthinkerthinkingstepbystepcompression, sui2025metareasonerdynamicguidanceoptimized}. Third, Reward-based and preference optimization methods, such as O1-Pruner \cite{luo2025o1prunerlengthharmonizingfinetuningo1like}, DAST \cite{shen2025dastdifficultyadaptiveslowthinkinglarge}, and IBPO \cite{xu2025chaindraftthinkingfaster}, leverage reinforcement learning or preference objectives to balance length and accuracy during generation \cite{qu2025optimizingtesttimecomputemeta}.

While effective, most existing methods apply static or globally optimized strategies, lacking adaptability to instance-specific reasoning complexity. We address this gap through multiround adaptive compression guided by token elasticity.

\section{Method}
\subsection{Token Elasticity Phenomenon} 
Recent studies have identified the \textit{Token Elasticity} phenomenon in LLMs \cite{han2025tokenbudgetawarellmreasoning}, where overly tight token budgets can lead to unexpected increases in output length due to compensatory and redundant generation. This reveals a nonlinear relationship between token constraints and actual model behavior. Motivated by this, we adopt multi-round progressive compression strategy that gradually tightens the CoT length over several steps. This allows the model to adapt more smoothly, avoiding abrupt information loss and mitigating the adverse effects of over-compression.

\subsection{CoT Generation}
Let $x$ denote the task input, and let $D_{\text{train}}$ be the training dataset and $\mathcal{P}$ be the prompt template. The initial CoT $C_0$ is generated by the target model $\mathcal{S}$ conditioned on $x$, $\mathcal{P}$, and parameters $\theta_{\mathcal{S}}$ learned from $D_{\text{train}}$:
\begin{equation}
r_0 = \mathcal{S}(x \mid \mathcal{P},\, \theta_{\mathcal{S}}(D_{\text{train}}))
\end{equation}

This initial CoT serves as the uncompressed sequence and is iteratively refined into shorter, semantically equivalent versions.

\subsection{Multi-Round Progressive Compression}

We then iteratively apply a sequence of $N$ compression process $\{f_1, f_2, \ldots, f_N\}$, implemented via an API-based compressor model, to produce a series of compressed CoTs:

\begin{equation}
r_i = f_i(r_{i-1} \mid \mathcal{P}_{\text{compress}}), \quad \text{for } i = 1, 2, \ldots, N
\end{equation}

Each $f_i$ operates over the previous CoT $C_{i-1}$ and is guided by a fixed compression prompt $\mathcal{P}_{\text{compress}}$. This design enables the gradual reduction of token length while attempting to preserve the correctness and reasoning validity of the original CoT.

To quantify the effect of compression at each stage, we define the compression rate at round $i$ as:

\begin{equation}
\mathrm{CR}_i=\frac{\left|r_i\right|_{\text {tok }}}{\left|r_0\right|_{\text {tok }}}
\end{equation}

where $\text {tok }$ denotes the number of tokens of the CoT sequence.

\par Our objective is to adaptively determine the maximum achievable compression rate for each input-specific CoT $r$, while preserving its reasoning validity. Instead of predefining a fixed target length or compression ratio, we propose a progressive compression framework that iteratively explores the compressibility of $r$ over multiple rounds. In each compression round $i$, a shorter variant $r_i$ is generated. The process terminates once the length of the newly generated chain $\mathrm{len}_{\text{tok}}(r_i)$ exceeds that of the previous round $\mathrm{len}_{\text{tok}}(r_{i-1})$, indicating that further compression leads to redundancy or loss of fidelity. In such cases, $r_{i-1}$ is selected as the maximally compressed yet valid chain $r^*$. Formally, the maximally compressed yet valid chain \( r^* \) is selected as:

\begin{equation}
\begin{aligned}
    r^* &= \arg\min_{r_j} \ \left| r_j \right|_{\text{tok}} \\
 &\text{subject to} \quad \left| r_j \right|_{\text{tok}} < \left| r_{j-1} \right|_{\text{tok}}
\end{aligned}
\end{equation}

where $\left| r_j \right|_{\text{tok}}$ denotes the tokenized length of $r_j$. The selected $r^*$ is subsequently used to fine-tune the target model. This adaptive criterion ensures compression proceeds only when meaningful token reduction is achieved, avoiding redundancy or semantic loss, and eliminating the need for manual compression targets. The process of the entire framework is shown in Algorithm \ref{alg:macc-dataset}. \footnote{For the reasoning model, we disentangle the CoT into the reasoning process and the answer process, apply compression to each component, and subsequently concatenate them to form the final representation.
}

\subsection{Multi-Task Fine-Tune}\label{multitask}

After obtaining the compressed rationale $r^*$ via the multi-round progressive compression framework described in Section 3.2, we employ a multi-task fine-tuning strategy to train the target model. We unify training on both original and compressed CoT by prepending a special token \texttt{<compress>}, denoted as $\boldsymbol{t_c}$ in the following format, signals the model to reason based on a concise chain of thought. Each training sample is thus formatted as:
$$
\mathcal{Q} \text { [EOS] } \boldsymbol{t_c} [\text{EOS}] \text { Compressed } \operatorname{CoT} r^*
$$

where $\langle\mathcal{Q}, \mathcal{A}\rangle$ indicates the $\langle\text{question}, \text{answer}\rangle$ pair. Formally, given a question $\boldsymbol{x}$, compression token $\boldsymbol{t_c}$, and the output sequence $\boldsymbol{y}=\left\{y_i\right\}_{i=1}^l$, which includes the compressed CoT $r^*$ and the answer $\boldsymbol{a}$, we fine-tunes the target LLM $\mathcal{S}$, enabling it to perform chain-of-thought in a compressed pattern by minimizing

\begin{equation}
\mathcal{L}=\sum_{i=1}^l \log P\left(y_i \mid \boldsymbol{x}, \boldsymbol{t_c}, \boldsymbol{y}_{<i} ; \boldsymbol{\theta}_{\mathcal{S}}\right)
\end{equation}

where $\boldsymbol{y}=\left\{c^*_1, \cdots, c^*_{m^{\prime}}, a_1, \cdots, a_t\right\}$. To retain the reasoning capabilities of LLMs, we include a fraction of original CoT trajectories in the training data, without setting $\boldsymbol{t_c}$.

\begin{algorithm}[t]
\caption{MACC: Multi-Round Adaptive Chain-of-Thought Compression for Dataset Construction}
\label{alg:macc-dataset}
\KwIn{Training set $D = \{x_j\}_{j=1}^N$, target model $\mathcal{S}$, compressor model $\mathcal{C}$, initial prompt $\mathcal{P}$, compression prompt $\mathcal{P}_{\text{compress}}$, max rounds $T$}
\KwOut{Compressed training set $D' = \{(x_j, r^*_j)\}_{j=1}^N$}

Initialize $D' \leftarrow \emptyset$\;

\ForEach{$x_j \in D$}{
    $r_0 \leftarrow \mathcal{S}(\mathcal{P}(x_j))$\;
    $r^* \leftarrow r_0$\;
    \For{$i = 1$ \KwTo $T$}{
        $r_i \leftarrow \mathcal{C}(\mathcal{P}_{\text{compress}}(x_j, r_{i-1}))$\;
        \If{$\mathrm{len}_{\text{tok}}(r_i) > \mathrm{len}_{\text{tok}}(r_{i-1})$}{
            \textbf{break}\;
        }
        $r^* \leftarrow r_i$\;
    }
    Add $(x_j, r^*)$ to $D'$\;
}
\Return{$D'$}
\end{algorithm}

\subsection{Inference}
MACC performs inference via autoregressive decoding. Given a question $x$ and a compression token $\boldsymbol{t_c}$, the input prompt follows the fine-tuning format: $\mathcal{Q} \ \text{[EOS]} \ \boldsymbol{t_c} \ \text{[EOS]}$. The LLM $\mathcal{S}$ then generates the output sequence $\hat{y}$ step by step:

$$
\hat{\boldsymbol{y}}=\arg \max _{\boldsymbol{y}^*} \sum_{j=1}^{l^{\prime}} \log P\left(y_j \mid \boldsymbol{x}, \boldsymbol{t_c}, \boldsymbol{y}_{<j} ; \boldsymbol{\theta}_{\mathcal{S}}\right)
$$
where $\hat{y} = \{\hat{c}_1, \dots, \hat{c}_{m''}, \hat{a}_1, \dots, \hat{a}_{t'}\}$ represents the generated output sequence, consisting of CoT tokens $\hat{c}$ and final answer tokens $\hat{a}$. The training and inference workflow of MACC is illustrated in Figure \ref{fig:detailed}.

\subsection{Performance Estimation Hypothesis}\label{pehypo}
Empirical observations suggest that the downstream performance of compressed CoT reasoning—measured by fine-tuned accuracy and CoT length—is correlated with interpretable features such as compression rate and perplexity. Based on this, we hypothesize that compressed performance can be predicted prior to fine-tuning:

\par \textit{Given a compressor model $\mathcal{C}$, a target model $\mathcal{S}$, and a training set $\mathcal{D}_{train}$, we define a feature vector $\boldsymbol{x}$ that encodes compression-related statistics, including the compression rate, perplexity, original CoT length, accuracy of both compressor and target model (the answer accuracy on training set). The downstream performance $\boldsymbol{y} = [\mathrm{Acc}, \mathrm{Len}]$ on test set $\mathcal{D}_{\text{test}}$ can be estimated as follows:}
\begin{equation}
P(\boldsymbol{y} \mid \boldsymbol{x}) = \frac{P(\boldsymbol{x} \mid \boldsymbol{y}) \cdot P(\boldsymbol{y})}{P(\boldsymbol{x})}
\label{eq:bayes}
\end{equation}

In practice, we approximate the posterior using Bayesian regression (e.g., Bayesian Ridge), yielding predictive distributions:
\begin{equation}
\boldsymbol{y}
\sim \mathcal{N}
\left(
\begin{bmatrix}
\mu_1 \\
\mu_2
\end{bmatrix},
\begin{bmatrix}
\sigma_1^2 & \rho \sigma_1 \sigma_2 \\
\rho \sigma_1 \sigma_2 & \sigma_2^2
\end{bmatrix}
\right)
\end{equation}
where $\mu(\cdot)$ and $\sigma^2(\cdot)$ denote the posterior mean and variance conditioned on the compression features.

This probabilistic formulation enables principled estimation of post-compression performance, supporting early-stage strategy selection without costly full fine-tuning. In practice, we implement this with Bayesian ridge regression, which provides both predictive means and uncertainty estimates. The effectiveness of this hypothesis is empirically validated in Section~\ref{sec:predictability}.

\section{Experiments}

\begin{table*}[ht!]
\centering
\resizebox{\textwidth}{!}{
\setlength{\tabcolsep}{4pt}
\renewcommand{\arraystretch}{1.1}
\begin{tabular}{llcccccccc}
\toprule
\multirow{2}{*}{\textbf{Methods}} & \multirow{2}{*}{\textbf{Model}} 
& \multicolumn{4}{c}{\textbf{GSM8K}} 
& \multicolumn{4}{c}{\textbf{MATH-500}} \\
\cmidrule(lr){3-6} \cmidrule(lr){7-10}
& & \makecell{Acc. $\uparrow$} & \makecell{Tokens $\downarrow$} 
& \makecell{Lat. (s) $\downarrow$} & \makecell{TE. $\uparrow$} 
& \makecell{Acc. $\uparrow$} & \makecell{Tokens $\downarrow$} 
& \makecell{Lat. (s) $\downarrow$} & \makecell{TE. $\uparrow$} \\
\midrule

\multirow{3}{*}{\textsc{Original}} 
& LLaMA-3.1-8B & 86.2 & 213.17 & 1.33 & 40.44 
& 48.6 & 502.60 & 6.83 & 9.67 \\
& Qwen2.5-7B   & 91.4 & 297.83 & 1.96 & 30.69 
& 71.4 & 574.85 & 6.65 & 12.42 \\
& Qwen2.5-3B   & 83.7 & 314.87 & 1.99 & 26.58 
& 61.6 & 578.51 & 5.90 & 10.65 \\
\midrule

\multirow{3}{*}{\textsc{Prompt}} 
& LLaMA-3.1-8B & 76.9 & 136.48 & 1.08 & 56.35 
& 37.6 & 335.92 & 3.78 & 11.19 \\
& Qwen2.5-7B   & 82.7 & 175.83 & 1.12 & 47.03 
& 49.1 & 355.47 & 3.45 & 13.81 \\
& Qwen2.5-3B   & 71.3 & 185.22 & 1.28 & 38.49 
& 42.0 & 423.88 & 3.98 & 9.91 \\
\midrule

\multirow{3}{*}{\textsc{TokenSkip}} 
& LLaMA-3.1-8B & 78.2 & 113.05 & 0.86 & 69.17
& 40.2 & 292.17 & 3.53 & 13.76 \\
& Qwen2.5-7B   & 86.0 & 151.44 & 0.89 & 56.79
& 52.8 & 330.8 & 3.12 & 15.96 \\
& Qwen2.5-3B   & 74.4 & \colorbox{cvprblue!14}{\textbf{170.55}} & \colorbox{cvprblue!14}{\textbf{1.02}} & \colorbox{cvprblue!14}{\textbf{43.62}}
& 44.2 & 396.29 & 3.74 & 11.15 \\
\midrule

\textsc{TALE} & LLaMA-3.1-8B & 78.5 & 139.63 & 0.88 & 56.22 & - & - & - & - \\
\midrule

\multirow{3}{*}{\textsc{MACC (Ours)}}
& LLaMA-3.1-8B & \colorbox{cvprblue!14}{\textbf{81.1}} & \colorbox{cvprblue!14}{\textbf{88.57}} & \colorbox{cvprblue!14}{\textbf{0.75}} & \colorbox{cvprblue!14}{\textbf{91.57}}
& \colorbox{cvprblue!14}{\textbf{44.0}} & \colorbox{cvprblue!14}{\textbf{198.04}} & \colorbox{cvprblue!14}{\textbf{2.05}} & \colorbox{cvprblue!14}{\textbf{22.22}} \\
& Qwen2.5-7B   & \colorbox{cvprblue!14}{\textbf{86.2}} & \colorbox{cvprblue!14}{\textbf{148.76}} & \colorbox{cvprblue!14}{\textbf{0.87}} & \colorbox{cvprblue!14}{\textbf{57.94}}
& \colorbox{cvprblue!14}{\textbf{58.4}} & \colorbox{cvprblue!14}{\textbf{254.89}} & \colorbox{cvprblue!14}{\textbf{2.02}} & \colorbox{cvprblue!14}{\textbf{22.91}} \\
& Qwen2.5-3B   & \colorbox{cvprblue!14}{\textbf{80.5}} & 216.25 & 1.33 & 37.22 
& \colorbox{cvprblue!14}{\textbf{54.0}} & \colorbox{cvprblue!14}{\textbf{265.80}} & \colorbox{cvprblue!14}{\textbf{2.20}} & \colorbox{cvprblue!14}{\textbf{20.32}} \\
\bottomrule
\end{tabular}
}
\caption{\textbf{Performance comparison on GSM8K and MATH-500 using three base models across five CoT compression methods}: Original, Prompt-only, TokenSkip, TALE, and our proposed MACC. Metrics include Accuracy, Token count, Latency (s), and Token Efficiency (Accuracy per token, scaled by 100). \colorbox{cvprblue!14}{\textbf{Bold}} values indicate the best results under each setting.}
\label{tab:main_results}
\end{table*}

\begin{table*}[ht!]
\centering
\resizebox{\textwidth}{!}{
\setlength{\tabcolsep}{4pt}
\renewcommand{\arraystretch}{1.1}
\begin{tabular}{llccccccccc}
\toprule
\multirow{2}{*}{\textbf{Methods}} & \multirow{2}{*}{\textbf{Model}} 
& \multicolumn{3}{c}{\textbf{GSM8K}} 
& \multicolumn{3}{c}{\textbf{MATH-500 (OOD)}} 
& \multicolumn{3}{c}{\textbf{AIME24 (OOD)}} \\
\cmidrule(lr){3-5} \cmidrule(lr){6-8} \cmidrule(lr){9-11}
& & \makecell{Acc. $\uparrow$} & \makecell{Tokens $\downarrow$} & \makecell{TE. $\uparrow$} 
  & \makecell{Acc. $\uparrow$} & \makecell{Tokens $\downarrow$} & \makecell{TE. $\uparrow$} 
  & \makecell{Acc. $\uparrow$} & \makecell{Tokens $\downarrow$} & \makecell{TE. $\uparrow$} \\
\midrule

\multirow{2}{*}{\textsc{Original}} 
& R1-Qwen-1.5B & 79.0 & 978  & 8.08  & 80.6 & 4887  & 1.65 & 29.4 & 12073 & 0.24 \\
& R1-Qwen-7B   & 87.9 & 682  & 12.89 & 90.2 & 3674  & 2.45 & 53.5 & 10306 & 0.52 \\
\midrule

\multirow{2}{*}{\textsc{O1-Pruner}} 
& R1-Qwen-1.5B & 74.8 & 458  & 16.33 & \colorbox{cvprblue!14}{\textbf{82.2}} & 3212  & 2.56 & \colorbox{cvprblue!14}{\textbf{28.9}} & 10361 & 0.28 \\
& R1-Qwen-7B   & 87.6 & 428  & 20.47 & 86.6 & 2534  & 3.42 & 49.2 & 9719  & 0.51 \\
\midrule

\multirow{2}{*}{\textsc{TALE}} 
& R1-Qwen-1.5B & 70.1 & 1170 & 5.99  & 76.2 & 3107  & 2.45 & 20.0 & 8915  & 0.22 \\
& R1-Qwen-7B   & \colorbox{cvprblue!14}{\textbf{91.0}} & 522  & 17.43 & \colorbox{cvprblue!14}{\textbf{91.6}} & 2530  & 3.62 & 33.3 & 8602  & 0.39 \\
\midrule

\multirow{2}{*}{\textsc{CoT Valve}} 
& R1-Qwen-1.5B & 70.4 & 805  & 8.74  & 76.5 & 2705  & 2.82 & 23.4 & \colorbox{cvprblue!14}{\textbf{5601}}  & \colorbox{cvprblue!14}{\textbf{0.42}} \\
& R1-Qwen-7B   & 90.8 & 364  & 24.94 & 89.4 & 1975  & 4.52 & 43.3 & 6315  & 0.69 \\
\midrule

\multirow{2}{*}{\textsc{MACC (Ours)}} 
& R1-Qwen-1.5B & \colorbox{cvprblue!14}{\textbf{79.3}} & \colorbox{cvprblue!14}{\textbf{471}}  & \colorbox{cvprblue!14}{\textbf{19.61}} & 76.0 & \colorbox{cvprblue!14}{\textbf{1954}}  & \colorbox{cvprblue!14}{\textbf{3.73}} & 26.7 & 7099  & 0.38 \\
& R1-Qwen-7B   & 90.1 & \colorbox{cvprblue!14}{\textbf{361}}  & \colorbox{cvprblue!14}{\textbf{27.37}} & 86.8 & \colorbox{cvprblue!14}{\textbf{2039}}  & \colorbox{cvprblue!14}{\textbf{5.74}} & \colorbox{cvprblue!14}{\textbf{50.0}} & \colorbox{cvprblue!14}{\textbf{6144}}  & \colorbox{cvprblue!14}{\textbf{0.81}} \\
\bottomrule
\end{tabular}
}
\caption{\textbf{Performance comparison of DeepSeek-R1-distill-Qwen-1.5B and DeepSeek-R1-distill-Qwen-7B on GSM8K, MATH-500 (OOD), and AIME24 (OOD) across five CoT compression methods}: Original, O1-Pruner, TALE and CoT-Valve. \colorbox{cvprblue!14}{\textbf{Bold}} values indicate the best results under each setting.}
\label{tab:deepseek_results}
\end{table*}

\subsection{Baseline}
To benchmark the effectiveness of our proposed compression framework, we compare against two recent and representative methods for efficient CoT reasoning:

\par $\bullet$ \textbf{TokenSkip} \cite{xia2025tokenskipcontrollablechainofthoughtcompression}. Compresses chain-of-thought by pruning low-importance tokens and fine-tuning the model to generate concise rationales based on a target compression ratio.

\par $\bullet$ \textbf{TALE} \cite{han2025tokenbudgetawarellmreasoning}. TALE controls CoT length by estimating token budgets from problem complexity, enabling efficient inference with minimal accuracy loss.

\par $\bullet$ \textbf{Prompt}. Following \citet{xia2025tokenskipcontrollablechainofthoughtcompression}, we guides the LLM to shorten its CoT output by incorporating explicit instructions into the prompt. For example, the input may include a directive such as: “Please reduce 50$\%$ of the words in your Chain-of-Thought reasoning.”

\par $\bullet$ \textbf{O1-Pruner} \cite{luo2025o1prunerlengthharmonizingfinetuningo1like}. O1-Pruner employs reinforcement learning–based fine-tuning to generate concise, non-redundant reasoning traces that preserve accuracy while enhancing efficiency and reducing computational cost.

\par $\bullet$ \textbf{CoT-Valve} \cite{ma-etal-2025-cot}. CoT-Valve learns a controllable parameter-space direction that adapts reasoning trace length to problem difficulty, reducing inference cost while maintaining competitive performance.

\subsection{Models and Datasets}
 For evaluation, we use three math datasets with increasing difficulty: GSM8K \cite{cobbe2021trainingverifierssolvemath}, MATH and AIME24 (30 Olympiad level math problems). For MATH, we evaluate on a 500-example subset (MATH-500) from \citet{lightman2023letsverifystepstep}, which reliably reflects full-benchmark performance. \texttt{GPT-4o-mini} serves as the compressor for its strong reasoning and efficiency \cite{openai2024GPT4omini}, while models include LLaMA-3.1-8B-Instruct \cite{grattafiori2024llama3herdmodels}, Qwen2.5-3B-Instruct and Qwen2.5-7B-Instruct \cite{yang2024qwen2technicalreport}, well as DeepSeek-R1-distill-Qwen-1.5B and DeepSeek-R1-distill-Qwen-7B \cite{deepseekai2025deepseekr1incentivizingreasoningcapability}, fine-tuned with compressed rationales via our multi-task strategy in Section \ref{multitask}. 

\subsection*{Evaluation Metrics}
We evaluate MACC using four key metrics to comprehensively assess reasoning performance and efficiency: accuracy (percentage of correctly answered questions), average CoT token count (to quantify reasoning verbosity), inference latency, and \textbf{Token Efficiency}.

We define Token Efficiency a composite metric defined as:
\[
\text{Token Efficiency} = \frac{Acc}{Length} \times 100
\]
where \(Acc\) denotes accuracy and \(Length\) the average CoT tokens. This metric captures the trade-off between accuracy and efficiency, with higher values indicating more effective reasoning and providing a comprehensive measure of performance and cost-effectiveness.

\subsection{Results}

\subsubsection{Main Result}
Table~\ref{tab:main_results} presents a comprehensive comparison of five CoT compression methods—Original, Prompt, TokenSkip, TALE, and our proposed \textbf{MACC}—on GSM8K and MATH-500, using three instruction-tuned models: LLaMA-3.1-8B-Instruct, Qwen2.5-7B-Instruct, and Qwen2.5-3B-Instruct. \textsc{MACC} consistently achieves the best trade-off between accuracy and efficiency across all settings.

On GSM8K, \textsc{MACC} improves accuracy over TokenSkip by +2.9 points on LLaMA-3.1-8B-Instruct and +0.2 points on Qwen2.5-7B-Instruct, while reducing the average CoT length by over 20\%. This leads to a clear gain in Token Efficiency, as fewer tokens are used without compromising performance. On MATH-500, which involves more complex reasoning, MACC continues to outperform TokenSkip, achieving +3.8 and +5.6 point gains in accuracy on LLaMA and Qwen2.5-7B, respectively. It also reduces inference latency (from 3.53s to 2.05s on LLaMA), highlighting its practicality for efficient reasoning. Even with the smaller Qwen2.5-3B model, MACC shows consistent improvements, confirming its robustness across different model capacities. In contrast, Prompt performs worse than TokenSkip in both accuracy and efficiency, leading to the lowest Token Efficiency overall. These results demonstrate that MACC effectively compresses CoT while preserving reasoning quality and reducing computational cost.

In Table \ref{tab:deepseek_results}, MACC demonstrates consistent advantages over existing CoT compression methods on both in-distribution (GSM8K) and out-of-distribution benchmarks (MATH-500, AIME24) across different reasoning models. Specifically, MACC achieves competitive or superior accuracy while substantially reducing the average reasoning length, leading to markedly higher token efficiency compared to O1-Pruner, TALE, and CoT-Valve. For example, on GSM8K, MACC attains comparable accuracy to the strongest baselines but with far shorter CoTs, yielding the highest token efficiency for both the 1.5B and 7B models. On the more challenging OOD datasets, MACC maintains stable accuracy while compressing reasoning traces more aggressively than competitors, thus striking a more favorable balance between efficiency and correctness. These results highlight the robustness and generalizability of MACC, showing that its adaptive multi-round compression strategy effectively mitigates performance degradation under distribution shift while delivering consistent efficiency gains.

\begin{figure}[!t]
\centering
\includegraphics[scale=0.22]{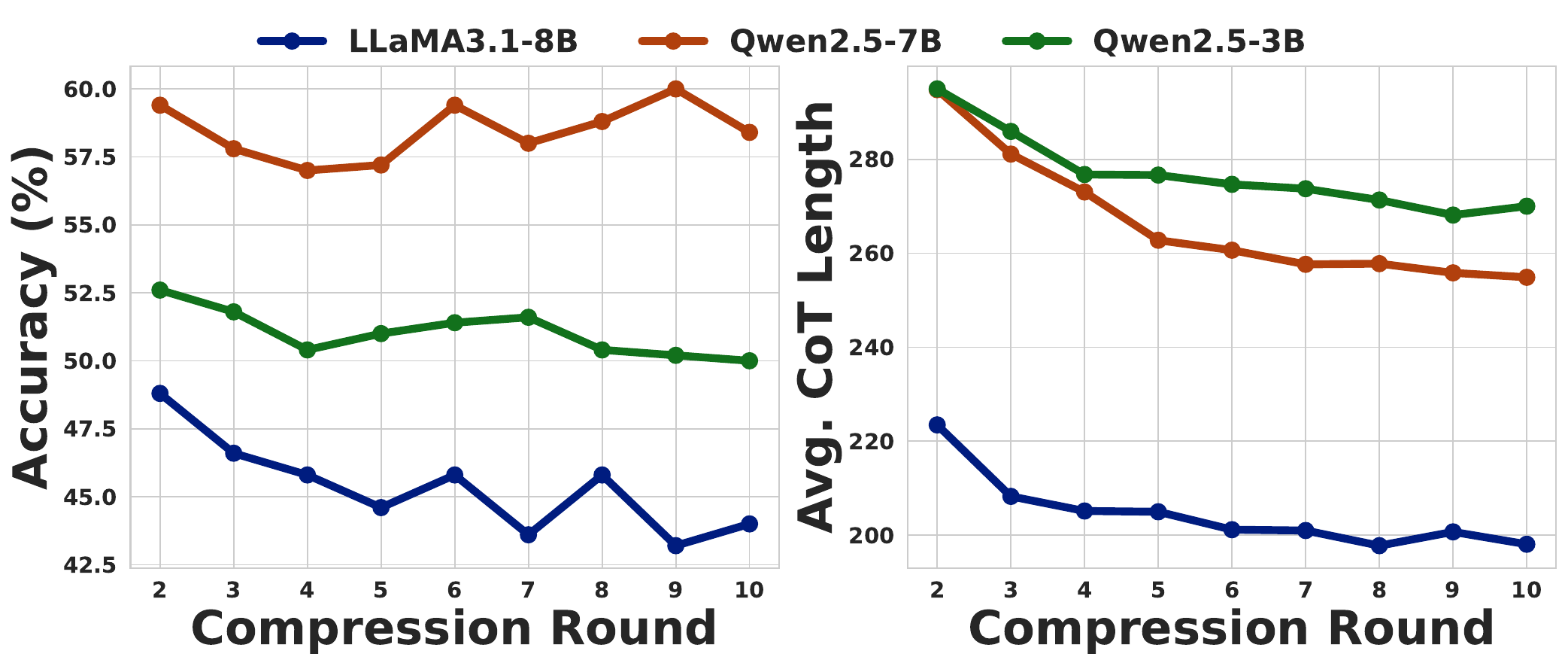}
\caption{
\textbf{Accuracy (left) and average CoT length (right) across compression rounds for three Models}. Larger models tend to retain higher accuracy under aggressive compression.}
\label{different_rounds}
\end{figure}

\begin{figure}[!t]
\centering
\includegraphics[scale=0.22]{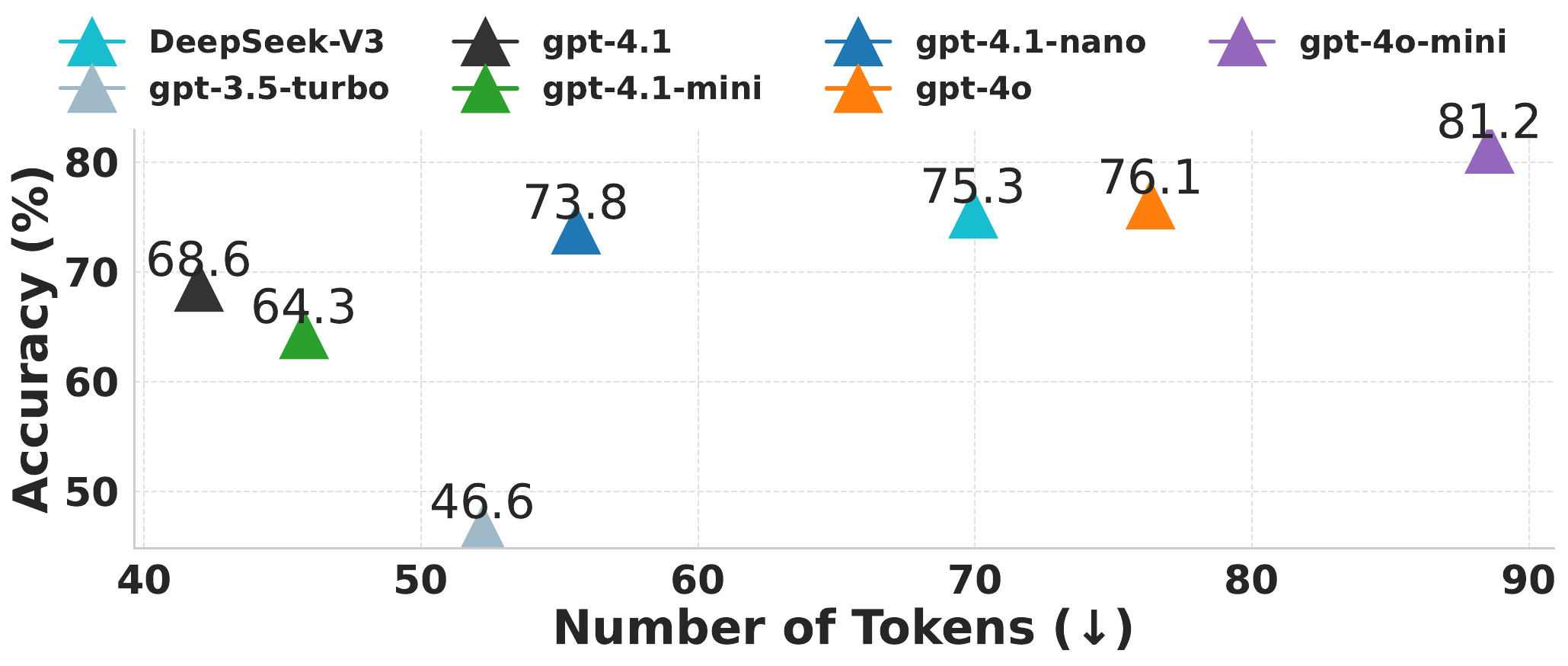}
\caption{\textbf{Comparison of compressed CoT performance across different compressors on llama3.1-8B-Instruct after 5 round}. Each point represents the accuracy and average CoT length achieved by different compressors. Models toward the upper-left indicate better trade-offs between efficiency and accuracy.}
\label{different_compressor}
\end{figure}

\subsubsection{Effect of Compression Rounds}
Figure~\ref{different_rounds} shows how fine-tuned accuracy and average CoT length evolve as the number of compression rounds increases under the MACC framework. As expected, the average length of the reasoning chains steadily decreases across rounds, demonstrating that MACC’s progressive strategy effectively eliminates redundant content while preserving the information necessary for reasoning.

The effect on accuracy, however, varies with model scale. Larger models like Qwen2.5-7B are more robust, maintaining high accuracy even with shorter CoTs. In contrast, smaller models suffer greater performance drops under aggressive compression, likely due to limited capacity to recover from incomplete rationales.

These results support the design of MACC’s adaptive stopping mechanism, which halts compression once further reduction harms accuracy. They also suggest that compression depth should be tailored to the model’s capacity, avoiding over-compression. Full results are provided in Table~\ref{tab:compression-llama3}, Table~\ref{tab:compression-qwen3b}, and Table~\ref{tab:compression-qwen7b} in Appendix~\ref{appA.4}.

\subsubsection{Effect of Different Compressor Models}
Next, we investigate how the choice of compressor model affects the quality of CoT compression, using LLaMA-3.1-8B-Instruct as the target model and GSM8K as the evaluation benchmark. As shown in Figure~\ref{different_compressor}, different compressors exhibit distinct trade-offs between compressed rationale length and fine-tuned accuracy across multiple compression rounds.

High-capacity compressors such as \texttt{GPT-4o} and \texttt{GPT-4o-mini} maintain high accuracy while significantly reducing CoT length, showing strong ability to preserve essential reasoning under compression. In contrast, lower-capacity models like \texttt{GPT-4.1-nano} and \texttt{GPT-3.5-turbo} cause greater accuracy drops, indicating weaker semantic fidelity and limited robustness in preserving logical consistency.

Overall, our results highlight that higher-capacity compressors tend to produce more compact yet informative rationales, enabling better fine-tuning outcomes. These findings underscore the importance of selecting an appropriate compressor model in multi-round compression pipelines, especially when targeting smaller or more sensitive student models.

\subsection{Estimating Compressed CoT Effectiveness}
Given the significant impact of compressor choice on the quality of CoT reasoning, it becomes increasingly important to assess, a priori, how a target model will perform when fine-tuned on compressed rationales. Instead of relying on exhaustive training and evaluation for every possible compression strategy, we investigate whether the downstream accuracy of the target model can be effectively predicted in advance. To this end, we explore a lightweight performance estimation framework conditioned on both the chosen compressor and the architecture of the target model. Specifically, we aim to estimate fine-tuned accuracy using a set of interpretable and readily available features extracted from compressed CoTs—such as token length, perplexity, and compression ratio. This approach enables efficient compression strategy selection without incurring the full cost of model retraining, and offers a practical pathway toward scalable and adaptive CoT compression.

Based on \textit{Performance Estimation Hypothesis}, we model their relationship in a probabilistic manner in Section \ref{sec:predictability}.


\begin{figure}[!t]
\centering
\includegraphics[scale=0.22]{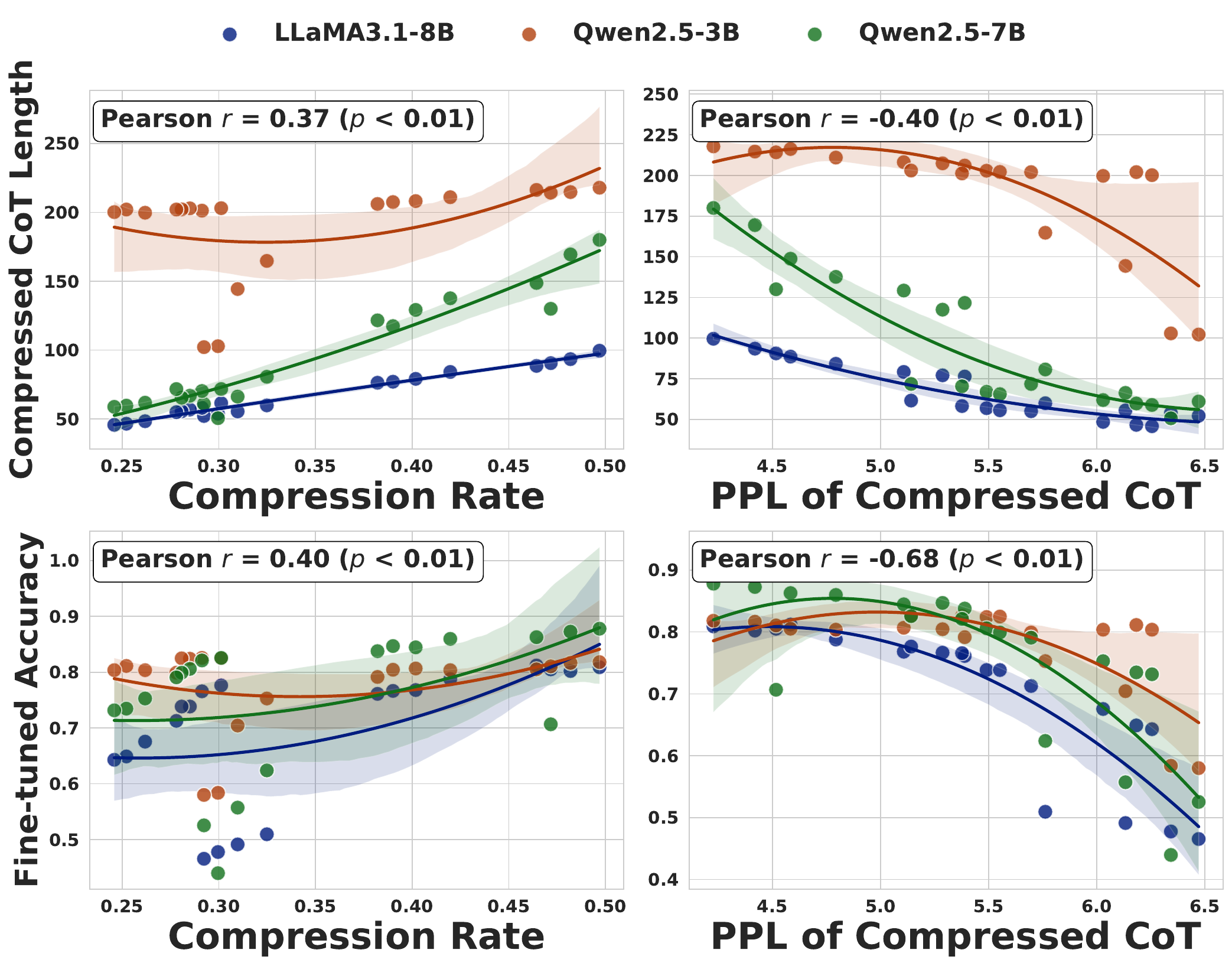}
\caption{\textbf{Effect of compression rate and perplexity on compressed CoT length and fine-tuned accuracy across different models.}. Each subplot shows the relationship between a compression feature and a target metric, with model-specific quadratic fits.}
\label{acc_correlation}
\end{figure}

\begin{figure}[!t]
\centering
\includegraphics[scale=0.22]{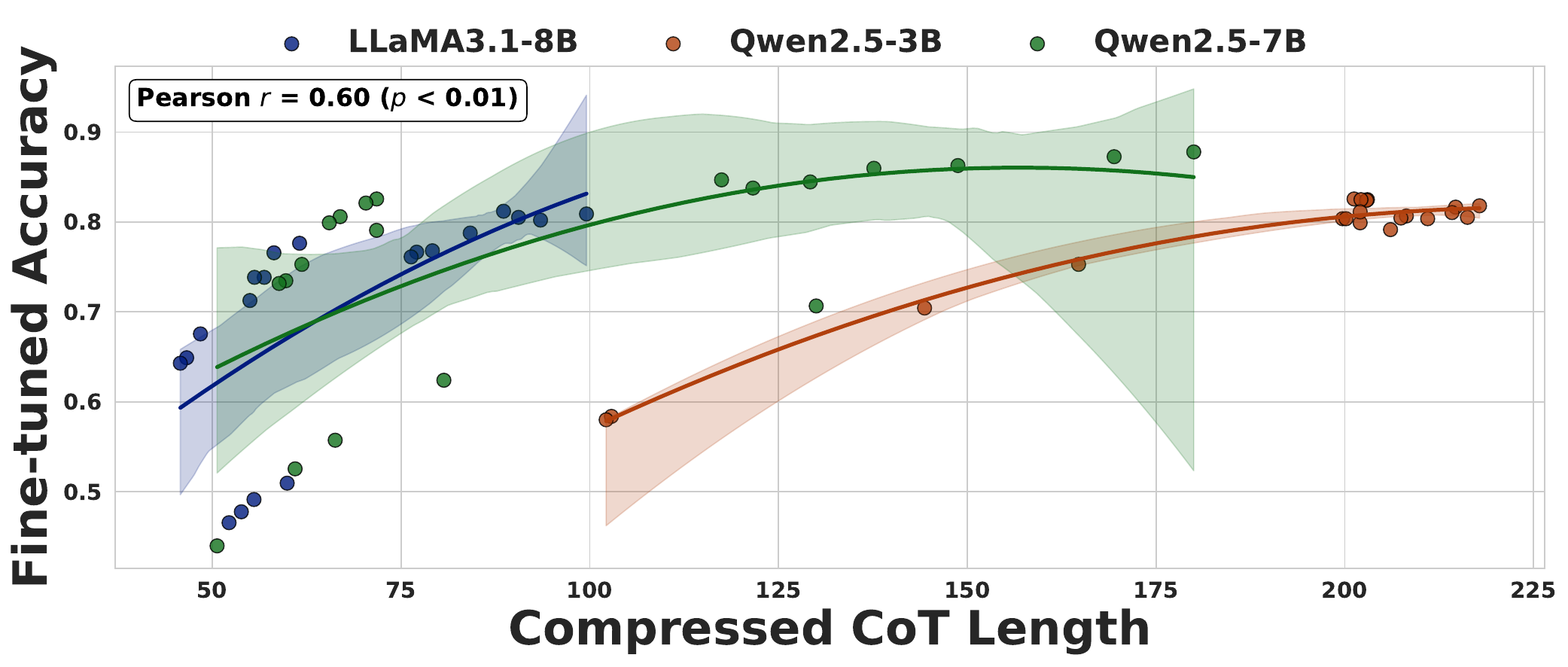}
\caption{\textbf{Relationship between compressed CoT length and fine-tuned accuracy across models}. Each point denotes a sample colored by target model. Longer compressed CoTs yield higher accuracy, suggesting the need to preserve key reasoning steps.}
\label{acc_length}
\end{figure}

\subsubsection{Analyse of Features}

To better understand the factors influencing post-compression performance, we analyze how interpretable features correlate with both the average CoT length and the fine-tuned accuracy. Figure~\ref{acc_correlation} show that compression rate and perplexity are moderately correlated with the resulting CoT length ($r = 0.37$ and $r = -0.40$, respectively), serving as a proxy for reasoning verbosity.

Figure \ref{acc_length} illustrates the relationship between the length of compressed CoT sequences and the downstream accuracy of fine-tuned models. A clear positive correlation is observed: longer compressed CoTs tend to yield higher fine-tuned accuracy. This highlights the need to preserve essential reasoning during compression, as over-truncation harms fidelity and performance. The results support the core design of the MACC framework, which adaptively determines compression depth to balance brevity and correctness. Additionally, the figure reveals that higher-capacity models achieve better accuracy at comparable CoT lengths, suggesting an interaction between model capacity and robustness to compression.

\subsubsection{Evaluating Predictability of Compressed CoT Effectiveness}\label{sec:predictability}

To validate the \textit{Performance Estimation Hypothesis}, we test whether fine-tuned accuracy can be predicted from interpretable features before training. We train regression models that take compression-related statistics—such as compression rate, perplexity of compressed CoT, original CoT length, and Compressor Accuracy—as inputs to estimate the downstream performance.

\begin{figure}[!t]
\centering
\includegraphics[scale=0.22]{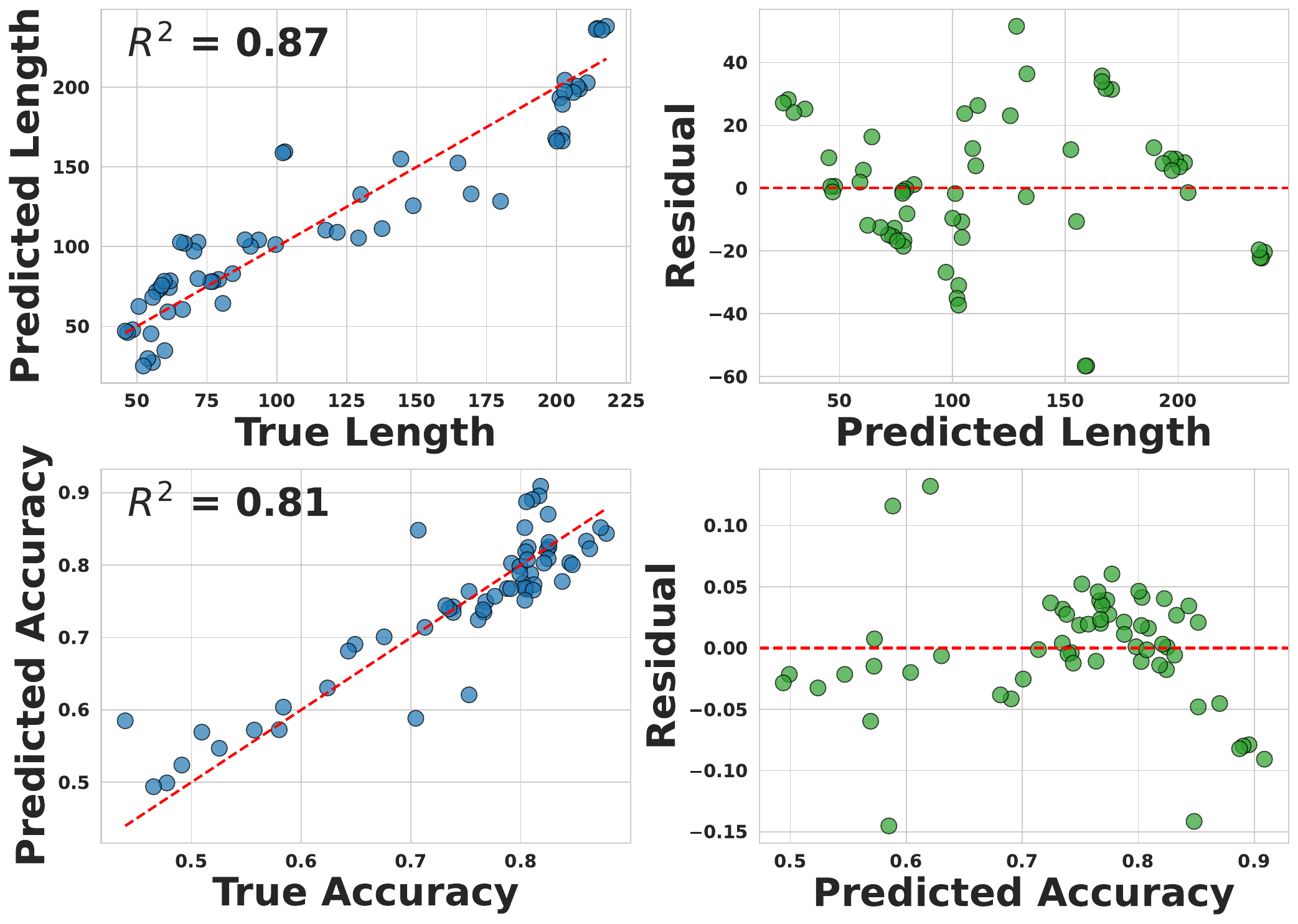}
\caption{\textbf{Bayesian Ridge regression results for predicting compressed CoT performance using features obtained from \textbf{training set}.}
Top row shows predictions and residuals for CoT length after compression; bottom row for fine-tuned accuracy. Predictions are based on training-set features before fine-tuning, demonstrating strong alignment with ground truth.}
\label{ridge_result}
\end{figure}

We experiment with both random forest and Bayesian ridge regressors under 5-fold cross-validation. As shown in Figure~\ref{ridge_result}, the predicted accuracies closely align with the true values, with low residuals across samples. This indicates that compressed CoT effectiveness is highly predictable using lightweight feature sets.

\begin{figure}[!t]
\centering
\includegraphics[scale=0.3]{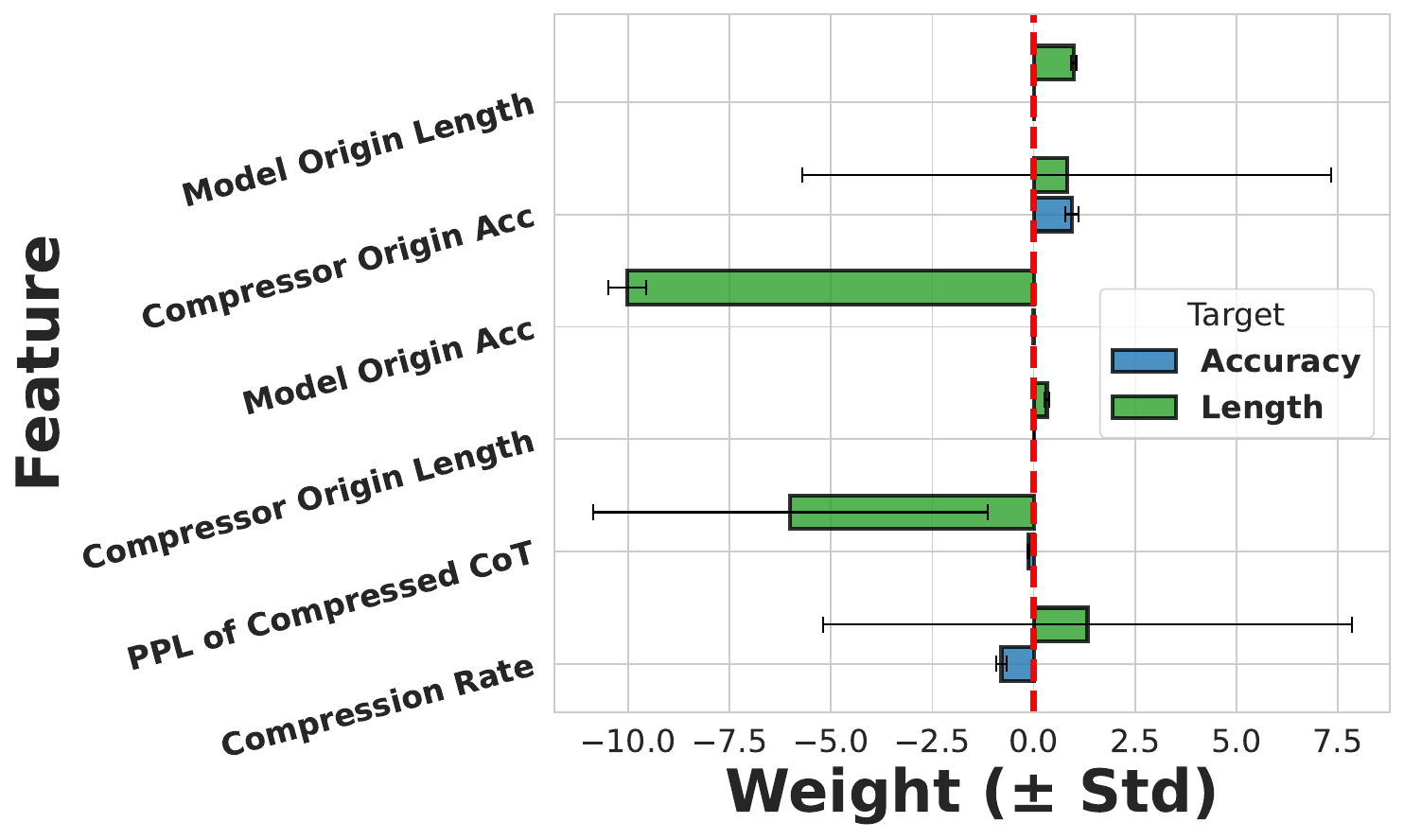}
\caption{\textbf{Bayesian Ridge regression weights for predicting fine-tuned accuracy and compressed CoT length using features extracted from the \textit{training set}}. Bars show feature importance (mean ± std), reflecting key factors for downstream performance.}
\label{feature_importance}
\end{figure}

Figure~\ref{feature_importance} shows that original CoT accuracy and length are the strongest predictors of fine-tuned accuracy, followed by compressed CoT perplexity. Notably, the compression rate itself contributes the least predictive signal, suggesting that surface-level reduction is less indicative of reasoning quality compared to semantic coherence or input-specific difficulty. These findings highlight the value of model- and CoT-aware features for estimating compression quality. 

These results support our hypothesis in Section~\ref{pehypo}: training-set features reliably predict performance, enabling efficient compressor selection without fine-tuning.

\section{Discussion}

\subsection{Performance Estimation as a Complement to MACC}
MACC’s adaptive stopping criterion optimizes compression depth but leaves compressor selection unresolved, as different compressors interact variably with the student’s knowledge distribution. To address this, the Performance Estimation Hypothesis (PEH) leverages interpretable features (e.g., perplexity, compression rate, accuracy) to predict compressed CoT effectiveness prior to fine-tuning, thereby enabling efficient compressor choice while MACC adaptively controls compression extent.

\subsection{Compressor as a Proxy for Target Model}
The effectiveness of CoT compression depends on both shortening and compressor–student compatibility; overly strong compressors may remove steps critical for smaller students, as seen when GPT-4o-mini yielded more suitable rationales than GPT-4o for mid-sized models.

Compression quality hinges on the alignment between compressor and student, and the Performance Estimation Hypothesis provides a lightweight means to assess this compatibility before fine-tuning, enabling efficient compressor–student selection.

\subsection{Contrasting CoT Compression and Distillation}
Although both MACC and CoT distillation exploit reasoning traces, their roles are distinct. Distillation transfers knowledge from a stronger teacher to a weaker student, enhancing reasoning capability, whereas MACC compresses self-generated CoTs via external APIs to optimize efficiency without introducing new knowledge. In essence, distillation focuses on ability transfer, while MACC emphasizes efficiency within existing capacity.

\section{Conclusion}
This paper presents \textbf{MACC}, a novel framework for adaptive and performance-aware compression of CoT reasoning. By leveraging the token elasticity phenomenon and multi-round refinement, MACC substantially reduces the length of reasoning chains with only minimal loss in accuracy. Extensive experiments across models and benchmarks demonstrate that MACC consistently outperforms prior approaches in terms of efficiency, accuracy, and latency. Furthermore, we show that key metrics such as post-compression accuracy and token usage can be reliably predicted using interpretable features. This enables informed compressor selection and efficient deployment, improving the scalability of CoT-based inference.

\clearpage
\section*{Limitations}
While MACC achieves substantial gains in compression efficiency and reasoning accuracy, it has several limitations. The reliance on external compressors (e.g., GPT-4o-mini) introduces potential model bias and limits applicability in low-resource settings. The multi-round process, while adaptive, adds preprocessing latency that may affect deployment speed. Additionally, compression prompts are task-agnostic, which may hinder performance on domains requiring structured reasoning. Lastly, our performance estimation relies on a limited feature set, which may not generalize well to unseen model-task combinations.

\section*{Acknowledgements}
This work is supported by the National Science and Technology Major Program  (2024ZD01NL00101), the Major Key Project of PCL (PCL2025A03), the Natural Science Foundation of China (62506182, 62276082), National Key RD Program of China (2023YFC3502900), Shenzhen Science and Technology Research and Development Fund (KJZD20240903102802003), Shenzhen Soft Science Research Program Project (RKX20220705152815035), Shenzhen Science and Technology Research and Development Fund for Sustainable Development Project (GXWD20231128103819001, KCXFZ20201221173613036, 20230706140548006) and Guangdong Provincial Key Laboratory (2023B1212060076).

\clearpage
\appendix

\section{Appendix}
\label{sec:appendix}

\subsection*{A.1 Correlation Analysis Between Compression Features and Downstream Performance}

Tables~\ref{tab:correlation_accuracy} and~\ref{tab:correlation_length} report the Pearson correlation coefficients between various compression-related features (e.g., perplexity, compression rate, original CoT length) and two key downstream metrics: accuracy and average CoT length after fine-tuning.

Specifically, Table~\ref{tab:correlation_accuracy} shows correlations with post-finetuning accuracy, while Table~\ref{tab:correlation_length} focuses on average CoT length. Across models, compressed perplexity and compression rate exhibit strong correlations with performance outcomes, validating their predictive utility. Notably, compressed perplexity tends to negatively correlate with accuracy and positively with output length, reinforcing its role as a proxy for semantic loss during compression. These results support the feasibility of estimating performance outcomes based on interpretable, compression-time statistics.
\begin{table}[h]
\centering
\footnotesize
\caption{Pearson correlation between compression-related features and fine-tuned accuracy.}
\label{tab:correlation_accuracy}
\resizebox{\linewidth}{!}{%
\begin{tabular}{lclS[table-format=1.3]S[table-format=1.1e-2]}
\toprule
\textbf{Model} & \textbf{Feature} & \textbf{$n$} &  \textbf{Pearson $r$} & \textbf{$p$-value} \\
\midrule
LLaMA-3.1-8B   & CR   & 20 &  0.540 & 1.3e-2  \\
LLaMA-3.1-8B   & PPL  & 20 & -0.810 & 1.3e-5  \\
LLaMA-3.1-8B   & Len  & 20 &  0.650 & 1.8e-3  \\
Qwen2.5-3B     & CR   & 20 &  0.210 & 3.9e-1  \\
Qwen2.5-3B     & PPL  & 20 & -0.560 & 1.1e-2  \\
Qwen2.5-3B     & Len  & 20 &  0.980 & 1.2e-14 \\
Qwen2.5-7B     & CR   & 20 &  0.420 & 6.5e-2  \\
Qwen2.5-7B     & PPL  & 20 & -0.700 & 6.5e-4  \\
Qwen2.5-7B     & Len  & 20 &  0.630 & 2.8e-3  \\
\bottomrule
\end{tabular}
}
\end{table}

\begin{table}[h]
\centering
\footnotesize
\caption{Pearson correlation between compression-related features and compressed CoT length.}
\label{tab:correlation_length}
\resizebox{\linewidth}{!}{%
\begin{tabular}{lclS[table-format=1.3]S[table-format=1.1e-2]}
\toprule
\textbf{Model} & \textbf{Feature} & \textbf{$n$} & \textbf{Pearson $r$} & \textbf{$p$-value} \\
\midrule
LLaMA-3.1-8B   & CR   & 20 &  0.990 & 5.5e-16 \\
LLaMA-3.1-8B   & PPL  & 20 & -0.930 & 1.9e-9  \\
LLaMA-3.1-8B   & Len  & 20 &  1.000 & 3.5e-20 \\
Qwen2.5-3B     & CR   & 20 &  0.370 & 1.1e-1  \\
Qwen2.5-3B     & PPL  & 20 & -0.670 & 1.3e-3  \\
Qwen2.5-3B     & Len  & 20 &  1.000 & 9.0e-21 \\
Qwen2.5-7B     & CR   & 20 &  0.960 & 5.0e-11 \\
Qwen2.5-7B     & PPL  & 20 & -0.910 & 3.6e-8  \\
Qwen2.5-7B     & Len  & 20 &  0.990 & 3.8e-17 \\
\bottomrule
\end{tabular}
}
\end{table}

\subsection*{A.2 Prompt Templates}

We provide the prompt templates used for both initial CoT generation and subsequent compression rounds.

\vspace{1mm}
\noindent\textbf{Initial CoT Generation Prompt:}
\begin{tcolorbox}[colback=gray!5!white, colframe=gray!50!black, boxrule=0.3pt]
\footnotesize
\texttt{
Please reason step by step, and put your final answer within \text{\\boxed{}}.\\
\\
QUESTION:\textbackslash n<Here is Question>\textbackslash n\\
}
\end{tcolorbox}

\vspace{2mm}
\noindent\textbf{Compression Prompt (for each round):}
\begin{tcolorbox}[colback=gray!5!white, colframe=gray!50!black, boxrule=0.3pt]
\footnotesize
\texttt{
You have a question now:\textbackslash n\\
QUESTION:\textbackslash n<Here is Question>\textbackslash n\\
THOUGHT PROCESS: <Here is Original CoT>\textbackslash n\\
ANSWER:\textbackslash n<Here is Final Answer>\textbackslash n\\
Now you need to simplify the THOUGHT PROCESS and retain the key information needed to solve the question.\\
And do not add additional information that is not included in the original THOUGHT PROCESS.\textbackslash n\\
SIMPLIFIED THOUGHT PROCESS:
}
\end{tcolorbox}

\subsection*{A.3 Example of Multi-Round CoT Compression}
Table \ref{tab:cot-multi-compression-clips} and \ref{tab:cot-multi-compression} presents a case study illustrating the progressive compression behavior of different compressors (GPT-4o-mini, GPT-3.5-turbo) on the same arithmetic question. Each row represents the reasoning trace generated at a specific compression round.

We observe that higher-capacity models (e.g., GPT-4o-mini) tend to produce more concise yet faithful compressed outputs in fewer rounds, while smaller models require more rounds to converge. Across all compressors, the CoT length gradually shrinks while preserving the correct final answer, which validates the effectiveness of our multi-round adaptive compression mechanism.

\begin{table*}[htbp]
  \centering
  \small
  \setlength{\tabcolsep}{6pt}
  \renewcommand{\arraystretch}{1.2}
  \begin{tabular}{c|>{\raggedright\arraybackslash}p{13.2cm}}
    \toprule
    \multicolumn{2}{>{\raggedright\arraybackslash}p{14.5cm}}{\textbf{Question:} Natalia sold clips to 48 of her friends in April, and then she sold half as many clips in May. How many clips did Natalia sell altogether in April and May?} \\
    \midrule

    \multirow{15}{*}{\textbf{GPT-4o-mini}} 
    & \textbf{Round 0 (Original):} \textit{Let's break it down step by step: 1. Natalia sold clips to 48 of her friends in April. 2. In May, she sold half as many clips as she did in April. To find half of 48, we divide 48 by 2: 48 ÷ 2 = 24. So, Natalia sold 24 clips in May. 3. To find the total number of clips she sold in April and May, we add the number of clips she sold in each month: 48 (April) + 24 (May) = 72. Therefore, Natalia sold 72 clips altogether in April and May. $\boxed{72}$}\\
    & \textbf{Round 1:} \textit{1. Natalia sold 48 clips in April. 2. In May, she sold half of that: 48 ÷ 2 = 24. 3. Total clips sold in April and May: 48 + 24 = 72. $\boxed{72}$} \\
    & \textbf{Round 2:} \textit{1. Natalia sold 48 clips in April. 2. In May, she sold 24 clips (half of April). 3. Total sold: 48 + 24 = 72. $\boxed{72}$} \\
    & \textcolor{RoyalBlue}{\textbf{Round 3:}} \textcolor{RoyalBlue}{\textit{1. Natalia sold 48 clips in April. 2. In May, she sold 24 clips (half of April). 3. Total sold: 48 + 24 = 72.}} \\
    & \textbf{Round 4:} \textit{1. Natalia sold 48 clips in April. 2. In May, she sold 24 clips (half of April). 3. Total sold: 48 + 24 = 72.} \\
    & \textbf{Round 5:} \textit{1. Natalia sold 48 clips in April. 2. In May, she sold 24 clips (half of April). 3. Total sold: 48 + 24 = 72.} \\
    \midrule

    \multirow{15}{*}{\textbf{GPT-3.5-turbo}} 
    & \textbf{Round 0 (Original):} \textit{Let's break it down step by step: 1. Natalia sold clips to 48 of her friends in April. 2. In May, she sold half as many clips as she did in April. To find half of 48, we divide 48 by 2:48 ÷ 2 = 24 3. So, Natalia sold 24 clips in May. 4. To find the total number of clips she sold in April and May, we add the number of clips she sold in each month: 48 (April) + 24 (May) = 72. Therefore, Natalia sold 72 clips altogether in April and May. $\boxed{72}$} \\
    & \textbf{Round 1:} \textit{Natalia sold 48 clips in April and half as many, which is 24 clips, in May, totaling $\boxed{72}$ clips sold altogether in April and May.} \\
    & \textcolor{RoyalBlue}{\textbf{Round 2:}}  \textcolor{RoyalBlue}{\textit{Natalia sold 48 clips in April and half as many, 24 clips, in May, totaling $\boxed{72}$ clips sold altogether in April and May.}} \\
    & \textbf{Round 3:} \textit{Natalia sold 48 clips in April and half as many, 24 clips, in May, totaling $\boxed{72}$ clips sold altogether in April and May.} \\
    & \textbf{Round 4:} \textit{Natalia sold 48 clips in April and half as many, 24 clips, in May, totaling $\boxed{72}$ clips sold altogether in April and May.} \\
    & \textbf{Round 5:} \textit{Natalia sold 48 clips in April and half as many, 24 clips, in May, totaling $\boxed{72}$ clips sold altogether in April and May.} \\
    \midrule

    \multirow{10}{*}{\textbf{DeepSeek-V3}} 
    & \textbf{Round 0 (Original):} \textit{Let's break it down step by step: 1. Natalia sold clips to 48 of her friends in April. 2. In May, she sold half as many clips as she did in April. To find half of 48, we divide 48 by 2:48 ÷ 2 = 24 3. So, Natalia sold 24 clips in May. 4. To find the total number of clips she sold in April and May, we add the number of clips she sold in each month: 48 (April) + 24 (May) = 72. Therefore, Natalia sold 72 clips altogether in April and May. $\boxed{72}$} \\
    & \textbf{Round 1:} \textit{1. April clips sold: 48 2. May clips sold: 48 ÷ 2 = 24. 3. Total clips sold: 48 + 24 = 72 $\boxed{72}$} \\
    & \textcolor{RoyalBlue}{\textbf{Round 2:}}  \textcolor{RoyalBlue}{\textit{1. April: 48 clips 2. May: Half of April → 24 clips 3. Total: $\boxed{72}$}} \\
    & \textbf{Round 3:} \textit{1. April: 48 clips 2. May: Half of April → 24 clips 3. Total: $\boxed{72}$} \\
    & \textbf{Round 4:} \textit{1. April: 48 clips 2. May: Half of April → 24 clips 3. Total: $\boxed{72}$} \\
    & \textbf{Round 5:} \textit{1. April: 48 clips 2. May: Half of April → 24 clips 3. Total: $\boxed{72}$} \\

    \midrule
    \multirow{15}{*}{\textbf{GPT-4o}} 
    & \textbf{Round 0 (Original):} \textit{Let's break it down step by step: 1. Natalia sold clips to 48 of her friends in April. 2. In May, she sold half as many clips as she did in April. To find half of 48, we divide 48 by 2:48 ÷ 2 = 24 3. So, Natalia sold 24 clips in May. 4. To find the total number of clips she sold in April and May, we add the number of clips she sold in each month: 48 (April) + 24 (May) = 72. Therefore, Natalia sold 72 clips altogether in April and May. $\boxed{72}$} \\
    & \textbf{Round 1:} \textit{1. Natalia sold 48 clips in April.  2. She sold half as many in May: (48 div 2 = 24).  3. Total clips sold in April and May: (48 + 24 = 72). $\boxed{72}$} \\
    & \textcolor{RoyalBlue}{\textbf{Round 2:}}  \textcolor{RoyalBlue}{\textit{Natalia sold 48 clips in April and half as many, 24 clips, in May, totaling $\boxed{72}$ clips sold altogether in April and May.}} \\
    & \textbf{Round 3:} \textit{Natalia sold 48 clips in April and half as many, 24 clips, in May, totaling $\boxed{72}$ clips sold altogether in April and May.} \\
    & \textbf{Round 4:} \textit{Natalia sold 48 clips in April and half as many, 24 clips, in May, totaling $\boxed{72}$ clips sold altogether in April and May.} \\
    
    \bottomrule
  \end{tabular}
  \caption{Example of Multi-Round CoT Compression results on \textbf{GSM8K} across different compressors. Each row shows the reasoning chain under different compression rounds. The selected CoT is marked in \textcolor{RoyalBlue}{blue}}
  \label{tab:cot-multi-compression-clips}
\end{table*}

\begin{table*}[htbp]
  \centering
  \small
  \setlength{\tabcolsep}{6pt}
  \renewcommand{\arraystretch}{1.2}
  \begin{tabular}{c|p{13.2cm}}
    \toprule
    \multicolumn{2}{l}{\textbf{Question:} What is the sum of the values of $x$ that satisfy the equation $x^2-5x+5=9$?} \\
    \midrule

    \multirow{70}{*}{\textbf{GPT-4o-mini}} 
    & \textbf{Round 0 (Original):} \textit{Step 1: Write down the given equation The given equation is $x^2-5x+5=9$.  Step 2: Subtract 9 from both sides of the equation to set it equal to zero Subtracting 9 from both sides gives $x^2-5x-4=0$.  Step 3: Use the quadratic formula to find the solutions for x The quadratic formula is $x=\frac{-b\pm\sqrt{b^2-4ac}}{2a}$, where $a=1$, $b=-5$, and $c=-4$.  Step 4: Plug the values of a, b, and c into the quadratic formula Plugging in the values gives $x=\frac{-(-5)\pm\sqrt{(-5)^2-4(1)(-4)}}{2(1)}$.  Step 5: Simplify the expression under the square root Simplifying the expression gives $x=\frac{5\pm\sqrt{25+16}}{2}$.  Step 6: Continue simplifying the expression under the square root Continuing to simplify gives $x=\frac{5\pm\sqrt{41}}{2}$.  Step 7: The sum of the values of x is the sum of the two solutions The sum of the values of x is $\frac{5+\sqrt{41}}{2}+\frac{5-\sqrt{41}}{2}$.  Step 8: Simplify the sum of the two solutions Simplifying the sum gives $\frac{10}{2}$.  Step 9: Simplify the fraction Simplifying the fraction gives $5$. The final answer is: $\boxed{5}$} \\
    \cmidrule(lr){2-2}
    & \textbf{Round 1:} \textit{Step 1: Write down the given equation The given equation is \( x^2 - 5x + 5 = 9 \).   Step 2: Set the equation to zero Subtracting 9 from both sides gives \( x^2 - 5x - 4 = 0 \).   Step 3: Use the quadratic formula Using the quadratic formula \( x = \frac{-b \pm \sqrt{b^2 - 4ac}}{2a} \) with \( a = 1 \), \( b = -5 \), and \( c = -4 \).   Step 4: Substitute values into the formula Substituting gives \( x = \frac{5 \pm \sqrt{(-5)^2 - 4(1)(-4)}}{2(1)} \).   Step 5: Simplify the expression This simplifies to \( x = \frac{5 \pm \sqrt{25 + 16}}{2} = \frac{5 \pm \sqrt{41}}{2} \).   Step 6: Find the sum of the solutions The sum of the solutions is \( \frac{5 + \sqrt{41}}{2} + \frac{5 - \sqrt{41}}{2} = \frac{10}{2} = 5 \).  The final answer is: \( \boxed{5} \)} \\
    \cmidrule(lr){2-2}
    
    & \textbf{Round 2:} \textit{Step 1: Write down the given equation The equation is  \( x^2 - 5x + 5 = 9  \).  Step 2: Set the equation to zero This leads to  \( x^2 - 5x - 4 = 0  \).  Step 3: Use the quadratic formula Using  \( x =  \frac{-b  \pm  \sqrt{b^2 - 4ac}}{2a}  \) with  \( a = 1  \),  \( b = -5  \),  \( c = -4  \).  Step 4: Substitute values into the formula This results in  \( x =  \frac{5  \pm  \sqrt{41}}{2}  \).  Step 5: Find the sum of the solutions The sum is  \(  \frac{5 +  \sqrt{41}}{2} +  \frac{5 -  \sqrt{41}}{2} = 5  \). Final answer:  \(  \boxed{5}  \)} \\
    \cmidrule(lr){2-2}
    & \textbf{Round 3:} \textit{Step 1: Write down the given equation The equation is  \( x^2 - 5x + 5 = 9  \).  Step 2: Set the equation to zero This leads to  \( x^2 - 5x - 4 = 0  \).  Step 3: Use the quadratic formula Using  \( x =  \frac{-b  \pm  \sqrt{b^2 - 4ac}}{2a}  \) with  \( a = 1  \),  \( b = -5  \),  \( c = -4  \).  Step 4: Substitute values into the formula This results in  \( x =  \frac{5  \pm  \sqrt{41}}{2}  \).  Step 5: Find the sum of the solutions The sum is  \(  \frac{5 +  \sqrt{41}}{2} +  \frac{5 -  \sqrt{41}}{2} = 5  \). Final answer:  \(  \boxed{5}  \)} \\
    \cmidrule(lr){2-2}
    & \textbf{Round 4:} \textit{Step 1: Write down the given equation The equation is \( x^2 - 5x + 5 = 9 \).  Step 2: Set the equation to zero This leads to \( x^2 - 5x - 4 = 0 \).  Step 3: Use the quadratic formula Using \( x = \frac{-b \pm \sqrt{b^2 - 4ac}}{2a} \) with \( a = 1 \), \( b = -5 \), \( c = -4 \).  Step 4: Substitute values into the formula This results in \( x = \frac{5 \pm \sqrt{41}}{2} \).  Step 5: Find the sum of the solutions The sum is \( 5 \). Final answer: \( \boxed{5} \)} \\
    \cmidrule(lr){2-2}
    & \textbf{Round 5:} \textit{Step 1: Write down the given equation The equation is \( x^2 - 5x + 5 = 9 \).  Step 2: Set the equation to zero This leads to \( x^2 - 5x - 4 = 0 \).  Step 3: Use the quadratic formula Using \( x = \frac{-b \pm \sqrt{b^2 - 4ac}}{2a} \) with \( a = 1 \), \( b = -5 \), \( c = -4 \).  Step 4: Substitute values into the formula This results in \( x = \frac{5 \pm \sqrt{41}}{2} \).  Step 5: Find the sum of the solutions The sum is \( 5 \). Final answer: \( \boxed{5} \)} \\
    \cmidrule(lr){2-2}
    & \textbf{Round 6:} \textit{Step 1: Write down the given equation The equation is \( x^2 - 5x + 5 = 9 \).  Step 2: Set the equation to zero This leads to \( x^2 - 5x - 4 = 0 \).  Step 3: Use the quadratic formula Using \( x = \frac{-b \pm \sqrt{b^2 - 4ac}}{2a} \) with \( a = 1 \), \( b = -5 \), \( c = -4 \).  Step 4: Substitute values into the formula This results in \( x = \frac{5 \pm \sqrt{41}}{2} \).  Step 5: Find the sum of the solutions The sum is \( 5 \). Final answer: \( \boxed{5} \)} \\
    \cmidrule(lr){2-2}
    & \textbf{Round 7:} \textit{Step 1: Write down the given equation The equation is \( x^2 - 5x + 5 = 9 \).  Step 2: Set the equation to zero This leads to \( x^2 - 5x - 4 = 0 \).  Step 3: Use the quadratic formula Using \( x = \frac{-b \pm \sqrt{b^2 - 4ac}}{2a} \) with \( a = 1 \), \( b = -5 \), \( c = -4 \).  Step 4: Substitute values into the formula This results in \( x = \frac{5 \pm \sqrt{41}}{2} \).  Step 5: Find the sum of the solutions The sum is \( 5 \). Final answer: \( \boxed{5} \)} \\
    \cmidrule(lr){2-2}
    & \textcolor{RoyalBlue}{\textbf{Round 8:}}  \textcolor{RoyalBlue}{\textit{Step 1: Write down the given equation The equation is \( x^2 - 5x + 5 = 9 \).  Step 2: Set the equation to zero Subtracting 9 gives \( x^2 - 5x - 4 = 0 \).  Step 3: Use the quadratic formula Using \( x = \frac{-b \pm \sqrt{b^2 - 4ac}}{2a} \) with \( a = 1 \), \( b = -5 \), \( c = -4 \).  Step 4: Substitute values into the formula This gives \( x = \frac{5 \pm \sqrt{25 + 16}}{2} = \frac{5 \pm \sqrt{41}}{2} \).  Step 5: Find the sum of the solutions The sum is \( \frac{5 + \sqrt{41}}{2} + \frac{5 - \sqrt{41}}{2} = \frac{10}{2} = 5 \).  The final answer is: \( \boxed{5} \)}} \\
    \cmidrule(lr){2-2}
    \bottomrule
  \end{tabular}
  \caption{Example of Multi-Round CoT Compression results on \textbf{MATH} across different compressors. Each row shows the reasoning chain under different compression rounds. The selected CoT is marked in \textcolor{RoyalBlue}{blue}}
  \label{tab:cot-multi-compression}
\end{table*}

\begin{figure}[!t]
\centering
\includegraphics[scale=0.22]{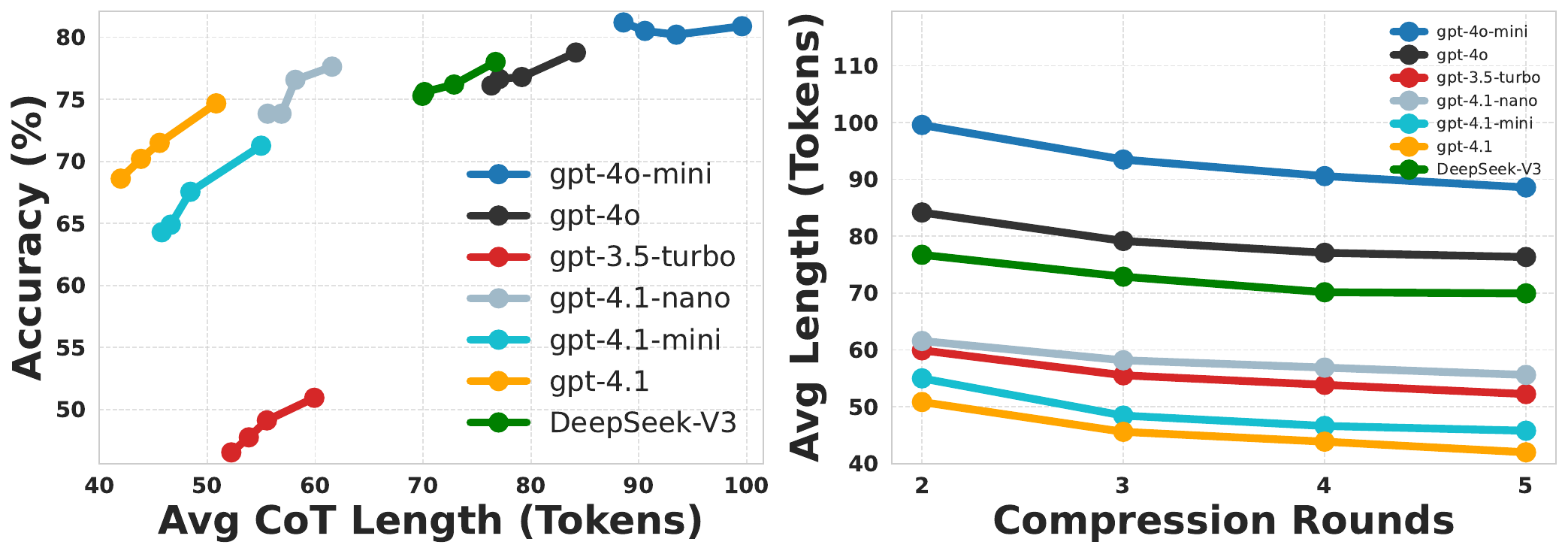}
\caption{Bayesian Ridge regression weights for predicting accuracy and CoT length using features obtained from \textbf{training set}. Bars show mean feature importance with standard deviation.
}
\end{figure}

\subsection*{A.4 Detailed Results of Compression and Fine-tuning}\label{appA.4}
Tables~\ref{tab:compression-llama3}, \ref{tab:compression-qwen3b}, and \ref{tab:compression-qwen7b} provide the full experimental results for all combinations of Model and compressor under different compression rounds. Each row shows the original Chain-of-Thought (CoT) length and accuracy, the compressed output’s perplexity and token length, and the downstream accuracy after fine-tuning the Model.

These detailed tables validate our key findings:
\begin{itemize}
  \item Increasing compression rounds leads to more compact reasoning traces but also higher perplexity.
  \item Compressors like GPT-4o and GPT-4o-mini consistently preserve semantic integrity better under aggressive compression, leading to superior fine-tuned accuracy.
  \item Lower-capacity compressors (e.g., GPT-3.5-turbo) experience sharper performance degradation under deeper compression.
\end{itemize}

We also observe that while the average CoT length drops by over 60\% in many cases, the fine-tuned accuracy retains over 90\% of its original value when using a well-matched compressor.

Table~\ref{tab:compression-llama3} reports the full results of multi-round compression and model fine-tuning across different compressors and Models. For each configuration, we list the number of compression rounds, the original CoT accuracy and length, the resulting compression rate, perplexity of the compressed CoT, and the fine-tuned model accuracy.

These results demonstrate the trade-off between compression depth and downstream performance. While deeper compression rounds reduce CoT length, they also tend to increase perplexity and reduce fine-tuned accuracy, especially under low-capacity compressors. Notably, models compressed by GPT-4o or GPT-4o-mini consistently outperform others in both efficiency and accuracy retention.

\subsection*{A.5 Implementation Details}
Table~\ref{tab:training-hparams} lists the hyperparameters used for Model fine-tuning across different datasets. We adopt LoRA \cite{hu2021loralowrankadaptationlarge}, an efficient and reproducible approach that has been widely verified as effective in LLM fine-tuning, to train our models. The rank $r$ is set to 8, and the scaling parameter $\alpha$ is set to 16. MACC is characterized by its low training cost, with training taking -1.5 hours for the 7B model. During inference, the maximum number of tokens is set to 16384. We implement our training process using the LLaMA-Factory \cite{zheng-etal-2024-llamafactory} library \footnote{\url{https://github.com/hiyouga/LLaMA-Factory}}.

\begin{table}[h]
\centering
\resizebox{\linewidth}{!}{%
\begin{tabular}{ll}
\toprule
\textbf{Parameter} & \textbf{Value} \\
\midrule
LoRA rank & 8 \\
LoRA alpha & 16 \\
Learning rate & $2 \times 10^{-5}$ \\
Batch size & 32 \\
Epochs & 3 \\
Max sequence length & 16384 \\
Precision & bfloat16 \\
Optimizer & AdamW \\
Scheduler & Cosine with warmup \\
\bottomrule
\end{tabular}
}
\caption{Fine-tuning hyperparameters for Models.}
\label{tab:training-hparams}
\end{table}

\subsection*{A.6 Performance Estimation Setup}

We use Bayesian Ridge regression as our default performance estimator. All features are normalized to zero mean and unit variance. We train one model per Model using 5-fold cross-validation with 80/20 train/test split.

For comparison, we also evaluate Random Forest regression with 100 trees, which shows similar but less interpretable results.

We report the average $R^2$ across folds for each model and target in Table~\ref{tab:prediction-results}.

\begin{table}[h]
\centering
\resizebox{\linewidth}{!}{%
\begin{tabular}{lcc}
\toprule
\textbf{Model} & $R^2$ (Accuracy) & $R^2$ (CoT Length) \\
\midrule
LLaMA3.1-8B & 0.81 & 0.87 \\
Qwen2.5-7B  & 0.78 & 0.91 \\
Qwen2.5-3B  & 0.73 & 0.89 \\
\bottomrule
\end{tabular}
}
\caption{Prediction performance of Bayesian Ridge on held-out data.}
\label{tab:prediction-results}
\end{table}

\begin{table*}[htbp]
\centering
\footnotesize
\caption{MACC compression results for Model: \textbf{LLaMA-3.1-8B}. Each row shows the result of multi-round compression using a specific compressor.}
\label{tab:compression-llama3}
\resizebox{\textwidth}{!}{%
\begin{tabular}{llcccccccc}
\toprule
\textbf{Model} & \textbf{Compressor} & \makecell{\textbf{Rounds}} & \makecell{\textbf{Original} \\ \textbf{Acc}} & \makecell{\textbf{Compressor} \\ \textbf{Acc}} & \makecell{\textbf{Original} \\ \textbf{Len}} & \makecell{\textbf{Compression} \\ \textbf{Rate}} & \textbf{PPL} & \makecell{\textbf{Compressed} \\ \textbf{Len}} & \makecell{\textbf{Finetuned} \\ \textbf{Acc}} \\
\midrule

LLaMA3.1-8B & GPT-3.5-turbo & 2 & 86.1 & 0.840 & 147.46 & 0.325 & 5.762 & 59.92 & 0.509 \\
LLaMA3.1-8B & GPT-3.5-turbo & 3 & 86.1 & 0.840 & 147.46 & 0.310 & 6.133 & 55.53 & 0.491 \\
LLaMA3.1-8B & GPT-3.5-turbo & 4 & 86.1 & 0.840 & 147.46 & 0.300 & 6.343 & 53.85 & 0.478 \\
LLaMA3.1-8B & GPT-3.5-turbo & 5 & 86.1 & 0.840 & 147.46 & 0.292 & 6.471 & 52.22 & 0.466 \\
LLaMA3.1-8B & GPT-4.1-mini  & 2 & 86.1 & 0.949 & 190.29 & 0.278 & 5.696 & 54.99 & 0.713 \\
LLaMA3.1-8B & GPT-4.1-mini  & 3 & 86.1 & 0.949 & 190.29 & 0.262 & 6.029 & 48.43 & 0.676 \\
LLaMA3.1-8B & GPT-4.1-mini  & 4 & 86.1 & 0.949 & 190.29 & 0.252 & 6.183 & 46.61 & 0.649 \\
LLaMA3.1-8B & GPT-4.1-mini  & 5 & 86.1 & 0.949 & 190.29 & 0.246 & 6.255 & 45.77 & 0.643 \\
LLaMA3.1-8B & GPT-4.1-nano  & 2 & 86.1 & 0.905 & 252.29 & 0.301 & 5.141 & 61.57 & 0.776 \\
LLaMA3.1-8B & GPT-4.1-nano  & 3 & 86.1 & 0.905 & 252.29 & 0.291 & 5.377 & 58.17 & 0.766 \\
LLaMA3.1-8B & GPT-4.1-nano  & 4 & 86.1 & 0.905 & 252.29 & 0.285 & 5.490 & 56.88 & 0.738 \\
LLaMA3.1-8B & GPT-4.1-nano  & 5 & 86.1 & 0.905 & 252.29 & 0.281 & 5.552 & 55.60 & 0.738 \\
LLaMA3.1-8B & GPT-4o        & 2 & 86.1 & 0.953 & 273.57 & 0.420 & 4.793 & 84.17 & 0.788 \\
LLaMA3.1-8B & GPT-4o        & 3 & 86.1 & 0.953 & 273.57 & 0.402 & 5.107 & 79.16 & 0.768 \\
LLaMA3.1-8B & GPT-4o        & 4 & 86.1 & 0.953 & 273.57 & 0.390 & 5.287 & 77.08 & 0.766 \\
LLaMA3.1-8B & GPT-4o        & 5 & 86.1 & 0.953 & 273.57 & 0.382 & 5.389 & 76.34 & 0.761 \\
LLaMA3.1-8B & GPT-4o-mini   & 2 & 86.1 & 0.922 & 330.37 & 0.497 & 4.227 & 99.57 & 0.809 \\
LLaMA3.1-8B & GPT-4o-mini   & 3 & 86.1 & 0.922 & 330.37 & 0.482 & 4.419 & 93.48 & 0.802 \\
LLaMA3.1-8B & GPT-4o-mini   & 4 & 86.1 & 0.922 & 330.37 & 0.472 & 4.517 & 90.57 & 0.805 \\
LLaMA3.1-8B & GPT-4o-mini   & 5 & 86.1 & 0.922 & 330.37 & 0.464 & 4.584 & 88.58 & 0.812 \\
\bottomrule
\end{tabular}
}
\end{table*}

\begin{table*}[htbp]
\centering
\footnotesize
\caption{MACC compression results for Model: \textbf{Qwen2.5-3B}.}
\label{tab:compression-qwen3b}
\resizebox{\textwidth}{!}{%
\begin{tabular}{llcccccccc}
\toprule
\textbf{Model} & \textbf{Compressor} & \makecell{\textbf{Rounds}} & \makecell{\textbf{Original} \\ \textbf{Acc}} & \makecell{\textbf{Compressor} \\ \textbf{Acc}} & \makecell{\textbf{Original} \\ \textbf{Len}} & \makecell{\textbf{Compression} \\ \textbf{Rate}} & \textbf{PPL} & \makecell{\textbf{Compressed} \\ \textbf{Len}} & \makecell{\textbf{Finetuned} \\ \textbf{Acc}} \\
\midrule
Qwen2.5-3B & GPT-3.5-turbo & 2 & 83.7 & 0.840 & 147.46 & 0.325 & 5.762 & 164.75 & 0.753 \\
Qwen2.5-3B & GPT-3.5-turbo & 3 & 83.7 & 0.840 & 147.46 & 0.310 & 6.133 & 144.35 & 0.704 \\
Qwen2.5-3B & GPT-3.5-turbo & 4 & 83.7 & 0.840 & 147.46 & 0.300 & 6.343 & 102.87 & 0.584 \\
Qwen2.5-3B & GPT-3.5-turbo & 5 & 83.7 & 0.840 & 147.46 & 0.292 & 6.471 & 102.17 & 0.580 \\
Qwen2.5-3B & GPT-4.1-mini  & 2 & 83.7 & 0.949 & 190.29 & 0.278 & 5.696 & 202.04 & 0.799 \\
Qwen2.5-3B & GPT-4.1-mini  & 3 & 83.7 & 0.949 & 190.29 & 0.262 & 6.029 & 199.71 & 0.804 \\
Qwen2.5-3B & GPT-4.1-mini  & 4 & 83.7 & 0.949 & 190.29 & 0.252 & 6.183 & 202.05 & 0.811 \\
Qwen2.5-3B & GPT-4.1-mini  & 5 & 83.7 & 0.949 & 190.29 & 0.246 & 6.255 & 200.17 & 0.804 \\
Qwen2.5-3B & GPT-4.1-nano  & 2 & 83.7 & 0.905 & 252.29 & 0.301 & 5.141 & 203.00 & 0.825 \\
Qwen2.5-3B & GPT-4.1-nano  & 3 & 83.7 & 0.905 & 252.29 & 0.291 & 5.377 & 201.23 & 0.826 \\
Qwen2.5-3B & GPT-4.1-nano  & 4 & 83.7 & 0.905 & 252.29 & 0.285 & 5.490 & 202.89 & 0.824 \\
Qwen2.5-3B & GPT-4.1-nano  & 5 & 83.7 & 0.905 & 252.29 & 0.281 & 5.552 & 202.15 & 0.825 \\
Qwen2.5-3B & GPT-4o        & 2 & 83.7 & 0.953 & 273.57 & 0.420 & 4.793 & 210.98 & 0.804 \\
Qwen2.5-3B & GPT-4o        & 3 & 83.7 & 0.953 & 273.57 & 0.402 & 5.107 & 208.16 & 0.807 \\
Qwen2.5-3B & GPT-4o        & 4 & 83.7 & 0.953 & 273.57 & 0.390 & 5.287 & 207.44 & 0.804 \\
Qwen2.5-3B & GPT-4o        & 5 & 83.7 & 0.953 & 273.57 & 0.382 & 5.389 & 206.06 & 0.792 \\
Qwen2.5-3B & GPT-4o-mini   & 2 & 83.7 & 0.922 & 330.37 & 0.497 & 4.227 & 217.85 & 0.818 \\
Qwen2.5-3B & GPT-4o-mini   & 3 & 83.7 & 0.922 & 330.37 & 0.482 & 4.419 & 214.69 & 0.817 \\
Qwen2.5-3B & GPT-4o-mini   & 4 & 83.7 & 0.922 & 330.37 & 0.472 & 4.517 & 214.21 & 0.810 \\
Qwen2.5-3B & GPT-4o-mini   & 5 & 83.7 & 0.922 & 330.37 & 0.464 & 4.584 & 216.25 & 0.805 \\
\bottomrule
\end{tabular}
}
\end{table*}

\begin{table*}[htbp]
\centering
\footnotesize
\caption{MACC compression results for Model: \textbf{Qwen2.5-7B}.}
\label{tab:compression-qwen7b}
\resizebox{\textwidth}{!}{%
\begin{tabular}{llcccccccc}
\toprule
\textbf{Model} & \textbf{Compressor} & \makecell{\textbf{Rounds}} & \makecell{\textbf{Original} \\ \textbf{Acc}} & \makecell{\textbf{Compressor} \\ \textbf{Acc}} & \makecell{\textbf{Original} \\ \textbf{Len}} & \makecell{\textbf{Compression} \\ \textbf{Rate}} & \textbf{PPL} & \makecell{\textbf{Compressed} \\ \textbf{Len}} & \makecell{\textbf{Finetuned} \\ \textbf{Acc}} \\
\midrule
Qwen2.5-7B & GPT-3.5-turbo & 2 & 91.4 & 0.840 & 147.46 & 0.325 & 5.762 & 80.68 & 0.624 \\
Qwen2.5-7B & GPT-3.5-turbo & 3 & 91.4 & 0.840 & 147.46 & 0.310 & 6.133 & 66.28 & 0.557 \\
Qwen2.5-7B & GPT-3.5-turbo & 4 & 91.4 & 0.840 & 147.46 & 0.300 & 6.343 & 50.64 & 0.440 \\
Qwen2.5-7B & GPT-3.5-turbo & 5 & 91.4 & 0.840 & 147.46 & 0.292 & 6.471 & 60.97 & 0.525 \\
Qwen2.5-7B & GPT-4.1-mini  & 2 & 91.4 & 0.949 & 190.29 & 0.278 & 5.696 & 71.76 & 0.791 \\
Qwen2.5-7B & GPT-4.1-mini  & 3 & 91.4 & 0.949 & 190.29 & 0.262 & 6.029 & 61.87 & 0.753 \\
Qwen2.5-7B & GPT-4.1-mini  & 4 & 91.4 & 0.949 & 190.29 & 0.252 & 6.183 & 59.74 & 0.735 \\
Qwen2.5-7B & GPT-4.1-mini  & 5 & 91.4 & 0.949 & 190.29 & 0.246 & 6.255 & 58.86 & 0.732 \\
Qwen2.5-7B & GPT-4.1-nano  & 2 & 91.4 & 0.905 & 252.29 & 0.301 & 5.141 & 71.77 & 0.826 \\
Qwen2.5-7B & GPT-4.1-nano  & 3 & 91.4 & 0.905 & 252.29 & 0.291 & 5.377 & 70.35 & 0.821 \\
Qwen2.5-7B & GPT-4.1-nano  & 4 & 91.4 & 0.905 & 252.29 & 0.285 & 5.490 & 66.94 & 0.806 \\
Qwen2.5-7B & GPT-4.1-nano  & 5 & 91.4 & 0.905 & 252.29 & 0.281 & 5.552 & 65.50 & 0.799 \\
Qwen2.5-7B & GPT-4o        & 2 & 91.4 & 0.953 & 273.57 & 0.420 & 4.793 & 137.63 & 0.860 \\
Qwen2.5-7B & GPT-4o        & 3 & 91.4 & 0.953 & 273.57 & 0.402 & 5.107 & 129.21 & 0.845 \\
Qwen2.5-7B & GPT-4o        & 4 & 91.4 & 0.953 & 273.57 & 0.390 & 5.287 & 117.46 & 0.847 \\
Qwen2.5-7B & GPT-4o        & 5 & 91.4 & 0.953 & 273.57 & 0.382 & 5.389 & 121.62 & 0.838 \\
Qwen2.5-7B & GPT-4o-mini   & 2 & 91.4 & 0.922 & 330.37 & 0.497 & 4.227 & 180.00 & 0.878 \\
Qwen2.5-7B & GPT-4o-mini   & 3 & 91.4 & 0.922 & 330.37 & 0.482 & 4.419 & 169.46 & 0.873 \\
Qwen2.5-7B & GPT-4o-mini   & 4 & 91.4 & 0.922 & 330.37 & 0.472 & 4.517 & 129.99 & 0.707 \\
Qwen2.5-7B & GPT-4o-mini   & 5 & 91.4 & 0.922 & 330.37 & 0.464 & 4.584 & 148.76 & 0.863 \\
\bottomrule
\end{tabular}
}
\end{table*}

\end{document}